\documentclass[11pt,a4paper]{article}
\usepackage{times,latexsym,amsfonts,amssymb}
\usepackage{hyperref}
\usepackage{url}
\usepackage{mathtools}
\usepackage[T1]{fontenc}
\usepackage{circuitikz}
\newcommand{\antonio}[1]
{{\color{orange!80!black}\textbf{[AF: #1]}}}

\usepackage{enumitem}
\setlist[itemize,enumerate]{leftmargin=\parindent,labelindent=\parindent,itemsep=1pt}

\usepackage{mdframed}
\definecolor{theoremcolor}{rgb}{0.94, 0.97, 1.0}
\definecolor{examplecolor}{rgb}{0.94, 0.97, 1.0}
\mdfsetup{
    innertopmargin=8pt,
    innerbottommargin=8pt,
    leftmargin=4pt,
    rightmargin=4pt,
    backgroundcolor=theoremcolor,
    linewidth=0pt,
}
\usepackage{xfrac}
\newmdtheoremenv[linewidth=0pt,innerleftmargin=4pt,innerrightmargin=4pt]{definition}{Definition}
\newmdtheoremenv[linewidth=0pt,innerleftmargin=4pt,innerrightmargin=4pt]{proposition}{Proposition}
\newmdtheoremenv[linewidth=0pt,innerleftmargin=0pt,innerrightmargin=0pt,backgroundcolor=examplecolor]{example}{Example}
\newmdtheoremenv{corollary}{Corollary}
\newmdtheoremenv{theorem}{Theorem}
\newmdtheoremenv{lemma}{Lemma}
\usepackage{comment}

\def\CP{CP\xspace}

\usepackage[acceptedWithA]{tacl2021v1}

\usepackage{xspace,mfirstuc,tabulary}

\newif\iftaclinstructions
\taclinstructionsfalse 
\iftaclinstructions
\renewcommand{\confidential}{}
\renewcommand{\anonsubtext}{(No author info supplied here, for consistency with
TACL-submission anonymization requirements)}
\newcommand{\instr}
\fi

\iftaclpubformat 

\else

\fi

\title{Conformal Prediction for Natural Language Processing: A Survey}

\usepackage{fontawesome5}
\newcommand{\lisbonit}[0]{$^1$}
\newcommand{\lisbonist}[0]{$^{,2}$}
\newcommand{\lumlis}[0]{$^{,3}$}
\newcommand{\unbabel}[0]{$^{,4}$}

\author{
  Margarida M. Campos \lisbonit\lisbonist \ 
  António Farinhas \lisbonit\lisbonist \
  Chrysoula Zerva \lisbonit\lisbonist\lumlis \
  \\
  \textbf{M\'ario A.T. Figueiredo}\lisbonit\lisbonist\lumlis \
  \and
   \textbf{Andr\'e F.T. Martins}\lisbonit\lisbonist\lumlis\unbabel
  \\
  \ \\
$^1$Instituto de Telecomunica\c{c}\~{o}es \
$^2$Instituto Superior T\'ecnico \\
$^3$LUMLIS (Lisbon ELLIS Unit)  \
$^4$Unbabel\\
   \texttt{margarida.campos@tecnico.ulisboa.pt}\\
}

\date{}

\begin{document}
\maketitle
\begin{abstract}
The rapid proliferation of large language models and natural language processing (NLP) applications creates a crucial need for uncertainty quantification to mitigate risks such as hallucinations and to enhance decision-making reliability in critical applications. Conformal prediction is emerging as a theoretically sound and practically useful framework, combining flexibility with strong statistical guarantees. Its model-agnostic and distribution-free nature makes it particularly promising to address the current shortcomings of NLP systems that stem from the absence of uncertainty quantification. This paper provides a comprehensive survey of conformal prediction techniques, their guarantees, and existing applications in NLP, pointing to directions for future research and open challenges. 
\end{abstract}

\section{Introduction}
Natural language processing (NLP) is witnessing an explosive growth in applications and public visibility, namely with large language models (LLMs) being deployed in many real-life applications, ranging from general-purpose chatbots to the generation of medical reports \citep{min2021recent}. 
However, the widespread use of these models brings important concerns: hallucinations are frequent \citep{ji2023survey, guerreiro2023hallucinations}, models are poorly calibrated  \citep{uncalibrated,desai-durrett-2020-calibration}, evaluation is limited and sometimes affected by data contamination \citep{contamination_1,contamination_2}, 
explanations are often unreliable 
\citep{zhao2023explainability,wiegreffe-pinter-2019-attention}, and models often exhibit undesired biases 
\citep{gallegos2024bias}. Reliable uncertainty quantification is key to addressing some of these concerns: NLP systems should not only provide accurate answers but also ``know when they do not know''. 

Unfortunately, most NLP systems return only single predictions (\textit{i.e.}, point estimates), without reliable confidence information.
Systems that quantify uncertainty are much less common and typically limited in various ways: they often make incorrect distribution-based assumptions ignoring the complex nature of the underlying data and model \cite{Xiao_Wang_2019,he-etal-2020-towards,glushkova-etal-2021-uncertainty-aware,zerva-etal-2022-disentangling}; they are often poorly calibrated (\textit{i.e.}, they predict a confidence level that does not match its error probability;  \citealt{pmlr-v80-kuleshov18a}); and they may be computationally too demanding, thus inapplicable to large-scale models \citep{hu2023uncertainty}.

\begin{figure*}
    \centering    \includegraphics[trim={4.6cm 1.6cm 2.6cm 3.4cm},clip,width=\textwidth]{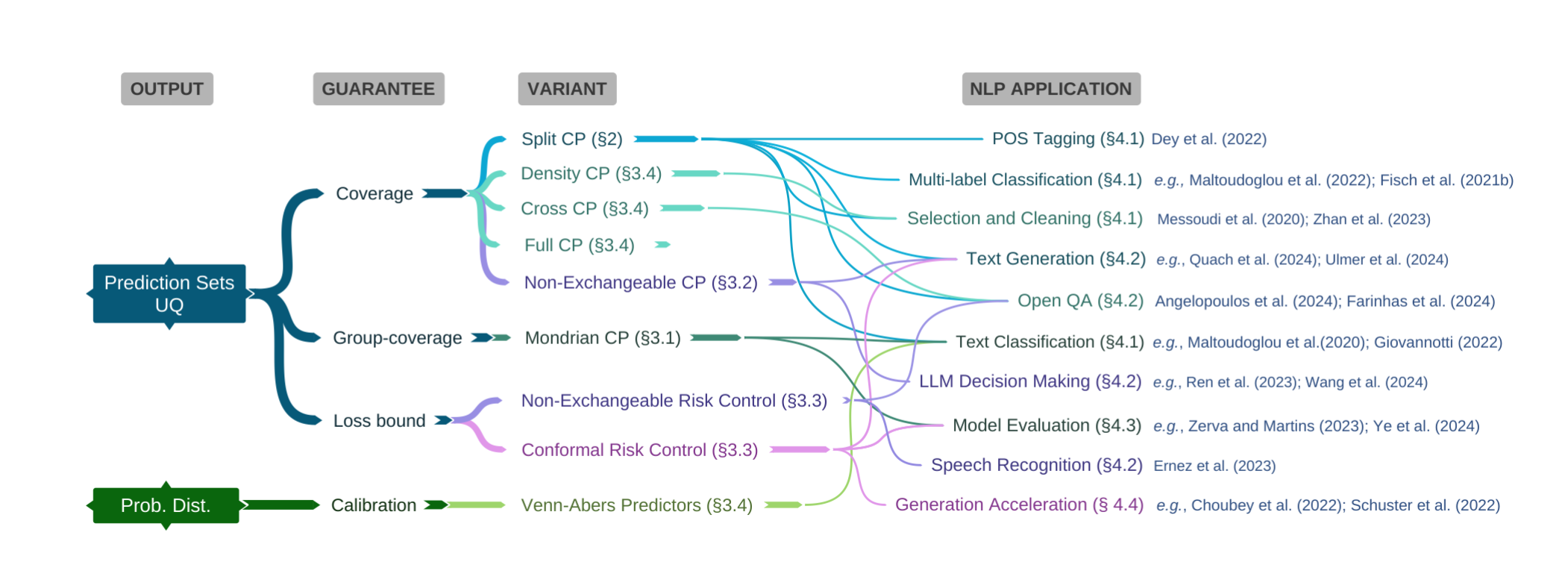}
    \caption{Survey roadmap: CP variants and their use in NLP applications with examples in the literature.}
    \label{fig:map}
\end{figure*}

\textbf{Conformal prediction} (CP; \citealt{vovk2005CP}) has recently emerged as a promising candidate to bypass the issues above: 
unlike other uncertainty quantification frameworks, it offers statistical guarantees of ground-truth coverage with minimal assumptions. \CP methods are \textbf{model-agnostic} and \textbf{distribution-free}, assuming only data exchangeability (as described in \S\ref{sec:guarantees}). Moreover, extensions of \CP that handle non-exchangeable data have recently been proposed \citep{gibbs2021adaptive, barber2023conformal}. Popular \CP variants are also \textbf{efficient}: they do not require model retraining and can be used online or offline, given an additional relatively small calibration set.\footnote{
For most purposes, a reasonable calibration set size is of the order of 1000 samples \citep{tutorial_gentle}.} 
Finally, equalized variants of \CP \citep{romano2019malice} can also reduce biases and unfairness, by distributing coverage evenly across protected attributes.  

The flexibility and strong statistical guarantees of \CP have attracted considerable interest, with an increasing number of publications in computer science.\footnote{The number of arXiv papers in the field of computer science containing the expression "conformal prediction" has been steadily rising, from \href{https://arxiv.org/search/advanced?advanced=&terms-0-operator=AND&terms-0-term=conformal+prediction&terms-0-field=all&classification-computer_science=y&classification-physics_archives=all&classification-include_cross_list=include&date-filter_by=specific_year&date-year=2018&date-from_date=&date-to_date=&date-date_type=submitted_date&abstracts=show&size=50&order=-submitted_date}{16 papers in 2018} to \href{https://arxiv.org/search/advanced?advanced=&terms-0-operator=AND&terms-0-term=conformal+prediction&terms-0-field=all&classification-computer_science=y&classification-physics_archives=all&classification-include_cross_list=include&date-filter_by=specific_year&date-year=2023&date-from_date=&date-to_date=&date-date_type=submitted_date&abstracts=show&size=50&order=-submitted_date}{224 in 2023}.} 
It is therefore timely to present a survey of conformal methods for NLP, revealing the theory and guarantees behind these methods and outlining opportunities and challenges for them to tackle important problems in the field.


\begin{figure*}
\centering
  \includegraphics[width=0.9\textwidth]{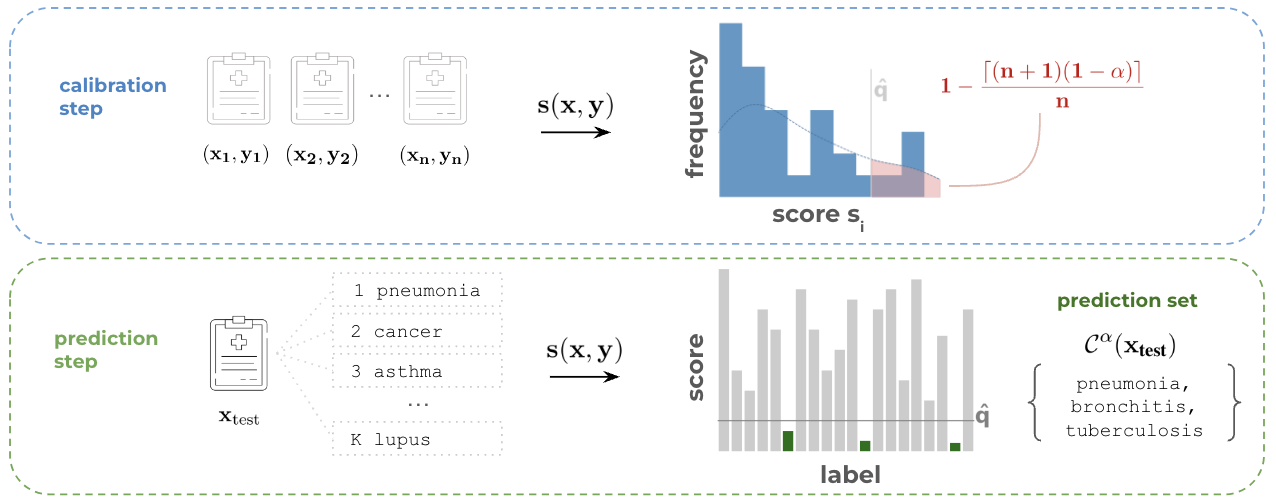}
  \caption{Example of CP for medical report classification ($K$ possible labels).}
  \label{fig:diagram}
\end{figure*}

\paragraph{Scope.}
This survey provides a comprehensive overview of \CP techniques for NLP tasks (Figure~\ref{fig:map}). After briefly explaining \CP and some relevant extensions (\S\ref{sec:Conformal Predictors} and \S\ref{sec:Extensions}), we review direct applications thereof in NLP (\S\ref{section:applications}). Finally, we look at possible threads of future investigation and current open issues concerning the use of \CP in NLP (\S\ref{sec:future}). 

\paragraph{What this survey is not about.}
This is \textit{not} a general survey on uncertainty quantification and does not include techniques not based on \CP$\!\!$.  Comprehensive reviews of uncertainty quantification in NLP were recently published by  \citet{baan2023uncertainty} and \citet{hu2023uncertainty}. Also, our survey is focused on NLP applications; \citet{tutorial_gentle} and \citet{tutorial_vovk} have published comprehensive surveys on \CP$\!\!$.


\section{Conformal Predictors}
\label{sec:Conformal Predictors}

This section briefly explains \CP and presents some definitions and results needed for understanding the applications mentioned below. 
In what follows, we use upper case letters ($X, Y, ...$) for random variables, lower case letters ($x, y, ...$)  for specific values they take, and calligraphic letters ($\mathcal{X}, \mathcal{Y}, \mathcal{C}, ...$) for sets.

\subsection{Definitions and Ingredients}\label{sec:defs}
Consider a prediction task where $\mathcal{X}$ and $\mathcal{Y}$ are the input and output sets, respectively. 
The most common procedure is to learn/train a mapping $f:\mathcal{X}\rightarrow \mathcal{Y}$, which, given an input $x_\mathrm{test} \in \mathcal{X}$, unseen during training, returns a \textbf{point prediction} $\hat{y}_\mathrm{test}=f(x_\mathrm{test}) \in \mathcal{Y}$, hopefully \textit{close} to the ``true" target $y_\mathrm{test}$, according to some performance metric.  A weakness of point predictions is the absence of information about uncertainty. In contrast, for the same input $x_\mathrm{test}$, a conformal predictor yields a \textbf{prediction set} $\mathcal{C_\alpha}(x_\mathrm{test})\subseteq \mathcal{Y}$, ideally small, which includes the target $y_\mathrm{test}$ with some high (user-chosen) probability, say $1-\alpha$. 

Consider an example involving a pretrained model which classifies a clinical report $x \in \mathcal{X}$ with a label, \textit{e.g.}, a disease $y \in \mathcal{Y}$. This is a high-risk  scenario requiring strong reliability guarantees. For a random test report ($X_\text{test}$), a conformal predictor yields a set $\mathcal{C}_\alpha(X_\text{test})$ of possibly multiples labels, with the guarantee that\footnote{Note that the probability is over $(X_\text{test}, Y_\text{test})$, \textit{not} conditioned on a particular $X_\text{test} = x_\text{test}$. We discuss conditional coverage in \S\ref{subsec:conditional}.}  $\mathbb{P}\left[Y_\text{test}\in\mathcal{C}_\alpha(X_\text{test})\right]\geq 1-\alpha$.
Figure \ref{fig:diagram} illustrates the \CP procedure for the mentioned task, which we describe next in detail.

Split\footnote{Although split (a.k.a. \textit{inductive}) \CP was developed after the \textit{full} (a.k.a. transductive) variant (described in \S\ref{subsec:other_extensions}), it is more widely used due to its computational efficiency.} \CP \citep{vovk2005CP} is built with three ingredients: a \textbf{trained predictor}, $f:\mathcal{X}\rightarrow \mathcal{Y}$; a \textbf{calibration set}, $\mathcal{D}_\text{cal} = \{(x_{1},y_{1}),...,(x_{n},y_{n})\}$, independent from the set used to train the predictor; and a \textbf{non-conformity score}, $s: \mathcal{X}\times\mathcal{Y}\rightarrow \mathbb{R}$. The non-conformity score measures how unlikely an input-output pair $(x,y) \in \mathcal{X}\times\mathcal{Y}$ is, compared to the remaining data. Consequently, given a test sample $x_\text{test}$, predictions $y\in \mathcal{Y}$ yielding pairs $(x_\text{test},y)$ deemed likely to occur in the data should have a low non-conformity score, and should thus be included in the prediction set $\mathcal{C}_\alpha(x_\mathrm{test})$. 

The choice of non-conformity score is task-dependent. For example, for a classifier outputting an estimate $p(y|x)$ of the posterior probability for each possible label $y\in\mathcal{Y}$ (\textit{e.g.}, via a softmax output layer), a common and natural choice is $s(x,y) = 1-p(y|x)$, with lower values of $s(x,y)$ implying that the sample is more conformal with the previously seen data.

\subsection{Procedure}\label{sec:procedure}
The procedure for generating $\mathcal{C}_\alpha(x_\text{test})$ for new, unseen test instances $x_\text{test}$ is as follows: 
\begin{enumerate}\setlength\itemsep{0em}
	\item Compute $(s_1, ... , s_n)$, the non-conformity scores for $\mathcal{D}_\mathrm{cal}$, where $s_i = s(x_i,y_i)$;
	\item Set $\hat{q}$ to be the $\lceil(n+1)(1-\alpha)\rceil/n$ empirical quantile of the set of scores;
	\item Output the prediction set, using the quantile $\hat{q}$, as $\mathcal{C}_\alpha(x_\text{test})=\{y\in\mathcal{Y}: s(x_\text{test},y)\leq \hat{q}\}$.
\end{enumerate}

Steps 1 and 2 are often referred to as \textbf{calibration}, and step 3 as \textbf{prediction}. The intuition is that the prediction set includes all predictions corresponding to samples that are more conformal than a sufficiently large fraction of the calibration set. 

\subsection{Theoretical Guarantees}
\label{sec:guarantees}
As shown by \citet{vovk2005CP}, a conformal predictor, as defined in the previous subsection, generates prediction sets with coverage guarantees, 
\begin{equation}
\mathbb{P}\left[Y_\text{test}\in\mathcal{C}_\alpha(X_\text{test})\right]\geq 1-\alpha,
	\label{eq:coverage}
\end{equation}
provided the data is exchangeable. Exchangeability means that the joint probability of the random variables generating the data is invariant under permutations thereof. Formally, a sequence $(Z_1,...,Z_n)$ is said to be \textbf{exchangeable} if
\begin{equation}
	\left(Z_1,...,Z_n\right) \overset{d}{=} \left(Z_{\pi(1)},...,Z_{\pi(n)}\right)
	\label{eq:exchangeability}
\end{equation}
for any permutations $\pi$ of $\{1,...,n\}$, where $\overset{d}{=}$ stands for \textit{identically distributed}. Exchangeability is a weaker requirement than the variables being independent and identically distributed (i.i.d). In fact, random variables that are i.i.d. are necessarily exchangeable; however, variables may be exchangeable without being independent, although they need to be 
identically distributed.
The coverage guarantee is provided by the following theorem \cite{vovk2005CP}: 
\begin{theorem}
	\label{theorem:icp}
	Let $(Z_1,...,Z_n,Z_\mathrm{test})$ be an exchangeable sequence of random variables, where $Z_i = (X_i,Y_i) \in \mathcal{X}\times\mathcal{Y}$, and $\mathcal{C}_\alpha: \mathcal{X}\rightarrow 2^\mathcal{Y}$ a conformal predictor as described in \S\ref{sec:procedure}. Then, $\mathcal{C}_\alpha$ satisfies
	\[
	1-\alpha\leq\mathbb{P}\left[Y_\text{test}\in\mathcal{C}_\alpha(X_\text{test})\right]\leq 1-\alpha+\frac{1}{n+1}.
	\]
\end{theorem}
 
A predictor satisfying the coverage inequality given in Theorem  \ref{theorem:icp} is said to be \textbf{valid}.%
\footnote{Altough there are other definitions of validity in the \CP literature \citep{vovk2005CP}, this is the most common one, termed \textit{conservative coverage validity}.} %
Note that as the size of the calibration set increases, the probability of coverage tends to exactly $1-\alpha$. It is worth noting that the \CP procedure we described is \textbf{model-agnostic} and \textbf{distribution-free}, \textit{i.e.}, it makes no assumption about the data distribution, requiring only data exchangeability.

\subsection{Relation to Hypothesis Testing} 
The \CP procedure described above can be seen from a hypothesis-testing perspective. For each possible label, the tested hypothesis is whether the point $(x_\mathrm{test},y)$ is conformal with the observed data, and the non-conformity measure is used as the test statistic. As an alternative to defining the threshold $\hat{q}$ using a preset $\alpha$, we can think in terms of empirical p-values \citep{vovk2005CP}. Define the p-value of a new sample $(x_{n+1},y_{n+1})$ as
\begin{align*}
\MoveEqLeft 
\text{p-value}(x_{n+1},y_{n+1}) = \\
& \hspace{-0.8cm}    \frac{\bigl| \{j\!\in\!\{1,..,n\}\! : s(x_j,y_j)\geq s(x_{n+1},y_{n+1})\}\bigr|+1}{n+1},
\end{align*}
the (adjusted) proportion of calibration points that are not less conformal than the observation. As in hypothesis testing, the p-value can be seen as the empirical probability of obtaining the observed score, under the null hypothesis that the observation is conformal. 
Using the p-value approach, the procedure to generate prediction sets for $x_\mathrm{test}$ is: 
\begin{enumerate}\setlength\itemsep{0em}
	\item compute p-values for all labels $y\in\mathcal{Y}$;
	\item generate prediction set as $\mathcal{C}_\alpha(x_\text{test})=\{y\in\mathcal{Y}: \text{p-value}(x_\mathrm{test},y)>\alpha \}$.
\end{enumerate}

A disadvantage of this approach is that it needs access to the calibration scores at test time. On the other hand, the p-values do not need a preset $\alpha$ and  can  be used to evaluate predictions, as shown next.

\subsection{Efficiency Metrics}
\label{subsec:evaluation}

When assessing the quality of a conformal predictor, an important aspect beyond validity is \textbf{efficiency}: the prediction sets should be relatively small and adaptive: easier cases should yield smaller sets than harder observations. The efficiency of a conformal predictor depends on the trained predictor $f$ and the chosen non-conformity score, which is typically based on some heuristic notion of prediction uncertainty, \textit{e.g.}, using the softmax output of a model (\S\ref{sec:defs}).  

Consider a separate test set $\mathcal{D}_\text{test} = \{(x_{n+1},y_{n+1}),...,(x_{n+k},y_{n+k})\}$. Some metrics, called \textit{a priori}, do not require access to the test set labels. This is the case of the \textbf{average prediction set size} (or interval width, in regression tasks): $
S(\alpha) = \frac{1}{k}\sum_{i=1}^{k} |\mathcal{C}_\alpha(x_{n+i})|$, computed as a function of $\alpha$.
Using the test set labels, an informative \textit{a posteriori} metric is the \textbf{observed fuzziness}, computed as the average of p-values for the false labels: $
\text{OF} = \frac{1}{k}\sum_{i=1}^k \sum_{y\neq y_{n+i}} \text{p-value}(x_{n+i},y)$,
which should be as small as possible, since correct predictions should
have high conformal scores, whereas incorrect labels should have low scores.
These metrics can also be useful to evaluate adaptivity and bias, by comparing them over different partitions of the dataset, \textit{e.g.}, split by a particular feature.
 
\subsection{Pointwise Metrics}
\label{sec:point-wise}
Conformal predictors provide point-level uncertainty metrics that can be used even in the \textbf{forced prediction} approach, \textit{i.e.,} producing as single prediction $\hat{y}_i$, the label with the highest p-value (typically coinciding with the original output of the point predictor), rather than the predicted set. Two common metrics in this case are \textbf{credibility}, $\mathrm{Cred}(x_i,\hat{y}_i) =  \text{p-value}(x_i,\hat{y}_i),$ and \textbf{confidence}, $\mathrm{Conf}(x_i,\hat{y}_i) = 1 - \underset{y\neq \hat{y}_i}{\max}\; \text{p-value}(x_i,y)$.
These metrics make use of the calibration set to measure uncertainty and can be extremely useful, even if disregarding the full prediction set produced by the conformal predictor.


\section{Extending Conformal Prediction}
\label{sec:Extensions}
\CP has extended beyond classic conformal predictors, with developments that allow handling challenges such as conditional coverage, dispensing with exchangeability, or obtaining guarantees beyond coverage. This section briefly presents the core ideas of some of the extensions that are most relevant for NLP applications. 

\subsection{Conditional Conformal Predictors}\label{subsec:conditional}
In many high-risk, critical settings, it may be important to obtain sample-conditional coverage,
\begin{equation}
\mathbb{P}\left[Y_\text{test}\in\mathcal{C}_\alpha(X_\text{test})|X_\text{test}=x_\text{test}\right]\geq 1-\alpha,
\label{eq:conditional_coverage }
\end{equation}
 for every $x_\text{test} \in \mathcal{X}$, \textit{i.e.,} provide a uniform upper bound for each prediction error. This, however, is not achievable under the proposed general setting, although, in practice, the error probability in some situations may be close to $\alpha$ \citep{vovk2012conditional, gibbs2023conformal, conditional_limits}. The study of conditional \CP is an active area of research with solutions to obtain coverage guarantees conditional on protected attributes and dataset partitions \citep{jin2024confidence,gibbs2023conformal,feldman2021improving}. This is extremely important for dealing with with class imbalance or fairness and bias concerns.

\citet{vovk2005CP}  introduced \textbf{Mondrian conformal predictors}:
conditional predictors that provide coverage guarantees over different data \textit{categories}, \textit{e.g.}: partitions of the data by label or by a given feature. For example, in classification, it may be of interest to have 
\begin{equation}
	\mathbb{P}\left[Y_\mathrm{test}\in \mathcal{C}_\alpha(X_\mathrm{test}) | Y_\mathrm{test}=y\right]\geq 1 - \alpha,
	\label{eq:label_cond}
\end{equation}
for all $y\in\mathcal{Y}$, which is a class-conditional guarantee. The procedure described in \S\ref{sec:procedure} is adapted to compute quantiles (or p-values) within each class. This is simply achieved by computing class-specific quantiles $\hat{q}^k$ based on the non-conformity scores of the calibration samples from each class $k$. Finally, the prediction set is given by 
\[
\mathcal{C}_\alpha(x_\mathrm{test})=\{y\in\mathcal{Y}: s(x_\mathrm{test},y)\leq \hat{q}^y\}.
\]

Assuming exchangeability (Eq. \ref{eq:exchangeability}), the above procedure is guaranteed to satisfy (Eq. \ref{eq:label_cond}). This label-conditional example is a particular case of Mondrian conformal predictors, which applies to any mapping of the data into Mondrian taxonomies \citep{vovk2005CP}. The same rationale can be used to obtain coverage across different partitions of the data, such as across a particular feature stratification.

\subsection{Beyond Exchangeability}
\label{subsec:CP beyond exchangeability}

All theoretical guarantees presented so far are rooted in the assumption of data exchangeability (Eq.~\ref{eq:exchangeability}). However, this assumption is unrealistic in many NLP applications: for example, it is incompatible with the conditional nature of most language generation methods. Several extensions have been proposed which handle \textbf{non-exchangeable data}, which includes the cases of covariate and label shift \citep{tibshirani2019covariate, podkopaev2021label}, time series \citep{chernozhukov2018exact, xu2021conformal, angelopoulos2023conformalpid}, and other types of shift \citep{gibbs2021adaptive}.

Recently, \citet{barber2023conformal} provided prediction guarantees without the exchangeability assumption. Let $Z_i = (X_i,Y_i) \in \mathcal{X}\times\mathcal{Y}$ be as defined in \S\ref{sec:guarantees}, $Z = (Z_1,...,Z_n,Z_{n+1})$ be a sequence of $n$ calibration samples followed by a test sample, and $Z^i$ denote $Z$ after swapping $Z_i$ with $Z_{n+1}$. \citet{barber2023conformal} proved that
\begin{equation}
	\mathbb{P}\left[Y_\text{test}\in\mathcal{C}_\alpha(X_\text{test})\right]\geq 1-\alpha \ - \sum_{i=1}^{n}\tilde{w}_i \, d_{\mathrm{TV}}(Z, Z^i),\label{eq:non-x-cp}
\end{equation}
where $\tilde{w}_i := w_i / (1 + \sum_{i=1}^N w_i)$  are weights (with $w_i \in [0, 1]$), and $d_{\mathrm{TV}}(Z, Z^i)$ is the {total-variation} distance between the distributions of $Z$ and $Z^i$. Choosing higher weights for calibration samples such that $Z$ and $Z^i$ have similar distributions yields tighter bounds.
Some open challenges related to this topic are discussed in \S\ref{sec:future}.

\subsection{Conformal Risk Control}
\label{subsec:Conformal Risk Control}
While coverage guarantees are useful in many tasks, there are cases where the adequate notion of \textbf{error control} is not captured solely by guaranteeing that the prediction set contains the ground truth. Some extensions of \CP address these cases. 

\citet{angelopoulos2024conformal} consider multilabel classification, where each $Y_i \in 2^\mathcal{Y} \setminus \{\varnothing\}$ is a set of labels. The loss function to be controlled is thus defined on pairs of sets of labels, $\ell \! :\! (2^\mathcal{Y} \setminus \{\varnothing\}) \times (2^\mathcal{Y} \setminus \{\varnothing\}) \! \! \rightarrow \mathbb{R}$, and assumed to satisfy \textbf{monotonicity}:  $A \subseteq B \Rightarrow \ell(A,Y) \geq \ell(B,Y)$, for any $Y \subseteq \mathcal{Y}$. They define prediction sets $\mathcal{C}_{\lambda}(x) = \{ y\in\mathcal{Y}: f(y|x) \geq 1 -\lambda\}$, where $f(y|x) \in [0,1]$ is the softmax output of class $y$, given by predictor $f$ for input $x$, and a parameter $\lambda$. Invoking loss monotonicity yields $\lambda \le \lambda' \Rightarrow \ell(\mathcal{C}_{\lambda},Y) \geq \ell(\mathcal{C}_{\lambda'},Y)$, for any $Y\subseteq\mathcal{Y}$. 

In this setting, and given some desired upper bound $\beta$ on the expected loss, \citet{angelopoulos2024conformal} propose a criterion to select a value $\hat{\lambda}$, such that the following bound holds: 
\begin{equation}
	\label{eq:conformal-risk-control}
	\mathbb{E}\Bigl[ \ell\bigl(\mathcal{C}_{\hat{\lambda}}(X_{\text{test}}),Y_{\text{test}}\bigr) \Bigr] \leq \beta.
\end{equation}
If $\ell$ is the miscoverage loss, \textit{i.e.}, $Y_{\text{test}}$ is a singleton and $\ell\bigl(\mathcal{C},Y_{\text{test}}\bigr) = 1 - | \mathcal{C} \cap  Y_{\text{test}}|$, the standard coverage guarantee in Eq.~\ref{eq:coverage} is recovered, with $\beta=\alpha$.

This is also related to (but different from)  previous work by \citet{bates2021rcps} and \citet{angelopoulos2022learn}, who prove  bounds of the form
\begin{equation}
    \mathbb{P}\left(\mathbb{E}\left[ \ell(Y_{\text{test}}, \mathcal{C}_{\hat{\lambda}}(X_{\text{test}})) \right] \leq \beta \right) \geq 1-\delta,
\end{equation}
where $\delta$ is a parameter and $\ell$ does not need to be monotone. \citet{angelopoulos2024conformal} provide comprehensive comparison of these so-called \textit{learn-then-test} (LTT) methods.
Finally, it is also possible to combine some of the ideas of \S\ref{subsec:CP beyond exchangeability} and \S\ref{subsec:Conformal Risk Control} to obtain non-exchangeable conformal risk control \citep{farinhas2024nonexchangeable}.


	
\subsection{Other CP Variants}
\label{subsec:other_extensions}
\paragraph{Full conformal prediction.} Introduced by \citet{vovk2005CP}, full \CP differs from the split version in two aspects: it does not use a separate calibration set, but the entire training set; and it involves model refitting---given a new instance, a model is trained for each possible label\footnote{For regression, discretization is typically used.} and used with the full data set to compute the non-conformity scores and obtain the prediction set. A clear disadvantage of full conformal prediction is the high computational cost of retraining. However, it has advantages: full conformal predictors can be used if there is a limited amount of data and model retraining is not too expensive, providing the same validity guarantees \citep{lei2017distributionfree}. 

\paragraph{Cross-validation and jackknifing.} The goal  of these methods is to achieve a balance between statistical and computational efficiency.  Cross-conformal predictors \citep{vovk2012crossconformal} apply the cross-validation rationale to split conformal predictors. Each cross-validation fold is used as a calibration set once and the p-values are computed using all folds. These predictors, although lacking proven validity guarantees, have shown good empirical results \citep{pmlr-v91-vovk18a}. 
Inspired by this idea, 
\citet{barber2020predictive} propose the so-called \textit{jackknife+}, a leave-one-out scheme, and prove validity for regression under some conditions.

\paragraph{Density-based conformal prediction.}
\citet{hechtlinger2019cautious} propose a different approach to the conformal procedure, based on $p(x|y)$ instead of the typical $p(y|x)$ to build more cautious predictors that should output the null set when underconfident. This method can be useful to abstain from answering when given an outlier observation. They show promising results using adversarial attacks on different tasks.

\paragraph{Venn-Abers predictors.} This class of probabilistic predictors has guarantees proved by \citet{vovk2014vennabers}. They produce one probability distribution per possible label and provide guarantees that one of the predictive distributions is perfectly calibrated, with no assumptions on the model or data distribution. Venn-Abers have been shown to be a good calibration tool with the added benefit that the distance between the different probability distributions provides calibrated uncertainty quantification \citep{johansson2023wellcalibrated}. A more efficient split variant is proposed by \citet{lambrou2014ivap}, and \citet{multi_class_ivap} presents a multi-class generalization.

\section{Applications in NLP} \label{section:applications}
\CP has been used in several NLP tasks, both to get validity/calibration guarantees on predictions; or within a pipeline, \textit{e.g.}: to safely prune intermediate outputs with guaranteed coverage, achieving computational speedups. This section reviews several such applications organized by use case.   

\subsection{Text Classification and  Sequence Tagging}


For classification and tagging tasks, models are often accurate but lack reliable confidence estimates. 

\paragraph{Binary text classification.} 
\citet{paisios2020bert} build a conformal predictor on top of a BERT classifier \cite{devlin2019bert} for binary sentiment classification. They show that the conformal predictor with forced prediction retains the original model's accuracy while providing useful accompanying measures of credibility and confidence. For the same task, \citet{messoudi2020deep} use density-based \CP (\S\ref{subsec:other_extensions}). They report good performance and empirical validity, highlighting the usefulness of having such a predictor by considering noisy and outlier observations: the \CP set contains both classes for the noisy example and is empty for the outliers, showing the desired discriminatory power. \citet{zhan2022reliable} automate identification of literature on drug-induced liver injury, using conformal prediction to manage prediction uncertainty and guaranteeing reliability.

\citet{giovannotti2022calibration} uses Venn-Abers predictors (\S\ref{subsec:other_extensions}) with different transformers model architectures on several binary tasks, such as paraphrase detection, sentiment analyses, and Boolean question answering, obtaining good calibration results with evenly distributed probability distributions.

\paragraph{Classification with conditional coverage.}
Mondrian \CP (\S\ref{subsec:conditional}) has been successfully applied to unbalanced classification tasks, such as sentiment analysis, with good efficiency results \cite{ulf2022sa}. \citet{giovannotti2021paraphrase} compare split, Mondrian and cross-conformal \cite{vovk2012crossconformal} \CP on  unbalanced paraphrase detection  
and report that the theoretically expected efficiency drop for Mondrian \CP is small,  making it useful in practice.

\paragraph{POS tagging.} \citet{dey2021conformal} present promising results by showing that \CP based on the softmax outputs of a BERT model for POS tagging yields practical prediction sets even at high confidence levels on a large test set: at the $99\%$ confidence level, fewer than $4\%$ of the prediction sets had more than one answer. 

\paragraph{Multilabel tasks.} \CP has been used for multilabel text classification, where multiple labels can be assigned to an input. In the label powerset approach \cite{Tsoumakas2010}, which treats each possible combination of labels as a class, there is an added challenge due to the large output space. \citet{paisios2019multilabel} show how \CP can be used in this setting, exploring different task-appropriate non-conformity scores. The forced prediction method (\S\ref{sec:point-wise}) shows negligible performance drops (as a consequence of part of the training data being set aside for calibration) while providing reliable credibility measures;  moreover, the prediction sets were tight and well-calibrated at high confidence levels. \citet{maltoudoglou2022multilabel} build on top of the aforementioned work and propose an efficient computational approach that allows a higher number of possible labels to be considered. \citet{fisch2022fp} tackle the multilabel case under the need to limit false positive predictions---a type of constraint that arises naturally in many highly sensitive tasks---by using a computationally efficient method that provides the desired coverage and constraint guarantees for an NER task, reporting prediction sets of useful size. 

A different approach has been considered in tasks such as document retrieval, where it may be of interest to obtain prediction sets with at least one admissible correct answer. \citet{fisch2021efficient} present an efficient conformal procedure to find such sets. They exploit the fact that simpler and lighter models can be used first in the pipeline to reduce the number of output candidates, producing a sequence of conformally valid candidates that are passed on to more complex models, showing that the final output is guaranteed to yield the user desired coverage. 

\paragraph{Dealing with limited data.}
\CP has also been found useful in providing guarantees for tasks with limited amounts of data. \citet{fisch2021fewshot} tackle few-shot  relation classification with \CP procedures to meta-learn both non-conformity measures and a threshold predictor from auxiliary tasks with larger amounts of available data. Not only do the predicted sets for the final task  grant coverage requirements, but they are also small (average set size smaller than 2 for 95\% confidence level). 
A different approach is used by \citet{dutta2023estimating} for estimating uncertainty in zero-shot biomedical image captioning using CLIP models \citep{radford2021learning}: they query the Web to get a calibration set and design a \CP protocol that takes into account the plausibility of each calibration point, providing promising results with small predicted test sets with coverage even in the absence of original labeled calibration data. In a setting with limited reliable data, \citet{zhan2023reliabilitybased} use \CP to clean possibly mislabeled training data, based on a small curated amount of data as a calibration set. They explore the effects of removing or changing the label of noisy data identified by the conformal procedure and show performance improvements on the text classification downstream task for different levels of induced noise. 

\subsection{Natural Language Generation}


Despite their impressive capabilities, large language models are prone to  {hallucinations} \citep{huang2023survey, ji2023survey}. The strong correlation between hallucinations and uncertainty unawareness makes \CP a promising approach to tackle this issue. However, its application to language generation faces two big challenges: (i) the combinatorially large size of output sets and (ii) the conditional (recursive) nature of language generation,  which violates the exchangeability assumption underlying standard \CP.

\paragraph{Sentence-level conformal prediction.}
Most research on \CP for 
NLP tries to circumvent the issues above by operating 
at the sentence level, \textit{e.g.}, 
by first sampling multiple options and then reformulating the problem as a multiple choice question \citep{kumar2023conformal}.
For instance, an LLM can be used to generate plans (expressed in natural language) for a robot to follow but a single plan alone may result in unfeasible or risky actions. 
\citet{ren2023robots} build upon the methods presented in \S\ref{sec:Conformal Predictors} to calibrate the confidence of \textbf{LLM planners}, providing formal guarantees for task completion while minimizing human help. Specifically, they look at the next-token probability to assess the uncertainty of different possible actions (\textit{i.e.}, they use it to compute the non-conformity score, as described in \S\ref{sec:procedure}) and generate \CP sets.
If the prediction set is not a singleton, the robot should ask for help; otherwise, it should continue to execute the plan. \citet{liang2024introspective} further enhance this framework by incorporating an {``introspective reasoning''} step \citep{Leake2012}, which leads to tighter prediction bounds, while \citet{wang2024safe} consider teams of robots.

\paragraph{Sentence-level risk control.}
\citet{quach2024conformal} show how LTT (\S\ref{subsec:Conformal Risk Control}) can be used to calibrate a stopping rule for sampling outputs from a language model that are added to a growing set of candidates until they are confident that the set includes at least one acceptable hypothesis \citep{fisch2021efficient}. Simultaneously, they calibrate a rejection rule to remove low-quality and redundant candidates. They use Pareto testing \citep{laufer-goldshtein2023efficiently} to efficiently search and test the high-dimensional hyperparameter configuration. The resulting output sets are not only valid 
but also precise (\textit{i.e.}, small). \citet{angelopoulos2024conformal} and \citet{farinhas2024nonexchangeable} apply conformal risk control to open-domain question answering, whereas \citet{ernez23applying} do it for speech recognition. While the former calibrate the best token-based $F_1$-score of the prediction set 
in Eq.~\ref{eq:conformal-risk-control}, the latter control the word error rate to an adjustable level of guarantee.
Finally, \citet{zollo2023prompt} discuss how prompts that perform well on average on a validation set may be prone to produce poor generations with high probability in deployment and propose \textbf{prompt risk control} based on upper bounds on families of informative risk measures.\footnote{They use the terms \emph{loss} and \emph{risk} in a distinctive way. Loss refers to scoring the quality of a single sample generation (\textit{e.g.}, ROUGE); risk measures some aspect of the distribution of the loss across the population (\textit{e.g.}, mean).} Specifically, they bound the worst-case toxicity \citep{Detoxify} in {chatbots}, the expected loss (pass@K, \citealt{kulal2019spoc}) in code generation, and the dispersion of ROUGE scores \citep{lin-2004-rouge} in medical summarization.

\paragraph{Token-level approaches.} While the approaches above focus on full sentences, language models generate text by successively producing new tokens autoregressively. 
Nucleus sampling \citep{Holtzman2020The} samples each token from the smallest set whose cumulative probability exceeds a  threshold. However, \citet{ravfogel-etal-2023-conformal} observe that LLMs  
tend to be overconfident---the prediction sets used in nucleus sampling are not calibrated (see their Fig.~4)---and this does not improve by scaling up the model size.
They propose \textbf{conformal nucleus sampling}, which calibrates prediction sets within bins of similar entropies. 
As an alternative, \citet{ulmer2024nonexchangeable} take non-exchangeability (\S\ref{subsec:CP beyond exchangeability}) into account by using a dynamic calibration step. 
They use the $k$-nearest neighbors and data-dependent {relevance} weights based on the squared $\ell_2$ distance between the embedding representations. This leads to smaller prediction sets compared to previous approaches while maintaining the desired coverage level in machine translation and language modeling. 

\subsection{Uncertainty-Based Evaluation}
\label{subsec:app_eval}

\CP can also be used to assist in evaluating and benchmarking NLP models. Two main approaches employ \CP to that end: (i) using it to assess the confidence of different models and compare them accordingly; (ii) framing evaluation as a regression task (\textit{i.e.}, learning to score the model outputs to predict human perceived quality and using \CP to provide reliable confidence intervals).  

Focusing on the former approach, \citet{ye2024benchmarking} apply \CP to benchmark the performance of different LLMs. They use prompt engineering to turn different generation tasks (question answering, summarization, commonsense inference, etc.) into multiple-choice questions such that the models need to predict a letter corresponding to each candidate output. 
They subsequently attempt to quantify the uncertainty of the language model over the possible labels, conformalizing the softmax outputs for each candidate label. They show that high model accuracy does not necessarily imply high certainty; in some cases, an inverse correlation between accuracy and certainty is observed. Based on their findings, \citet{ye2024benchmarking} propose an uncertainty-aware metric accounting for both accuracy and uncertainty (encoded as set size).



Focusing instead on the latter approach, \citet{giovannotti2023evaluating} applies \CP to {referenceless MT evaluation} (quality estimation) and uses a $k$-nearest neighbor model to obtain quality scores and subsequently use the distances between each point and its neighbors to form non-conformity scores. They thus use \CP as a method to quantify uncertainty for MT quality estimation. \citet{zerva2023conformalizing}, on the other hand, apply \CP on top of non-conformity heuristics coming from other uncertainty quantification methods for reference-based MT evaluation and discuss how such method choice can impact coverage and width. They also highlight biases in estimated confidence intervals, reflected in imbalanced coverage for attributes such as translation language and quality, demonstrating how these can be addressed with equalized \CP. 
While focused on MT, the proposed approaches are applicable to other NLP evaluation or regression tasks. 


\subsection{Faster Inference} 

Given the high computational requirements of state-of-the-art NLP models and their widespread use, considerable effort is being put on making these models more time- and memory-efficient  \cite{survey_acceleration}. Several strategies for increasing efficiency at prediction time (\textit{e.g.}, early exiting,  \citealt{liu2019earlyexit, schwartz2020earlyexit}) focus on identifying easily classifiable instances and using a lighter version of the original model to predict them. Such instances must be reliably identified and both the original and simplified models should consistently produce the same results for a given input with a high probability.

\paragraph{Early exiting transformers fine-tuning.} \citet{schuster2021consistent} present an extension of \CP to build a method to speed up inference in transformer models, while guaranteeing an  adjustable degree of consistency with the original model, with high confidence. The rationale is to skip directly to the final layer from one of the previous layers  whenever there is enough confidence. They use a binary meta-classifier to predict whether the lighter model is consistent with the original one and use \CP to predict the set of inconsistent models. The final procedure consists of exiting at the first layer that exceeds the threshold found by the conformal procedure. Their method shows reduced inference time in several classification and regression tasks.

\paragraph{Zero-shot learning.} \citet{choubey-etal-2022-conformal} tackle the computational efficiency problem in zero-shot 
text classification with pretrained language models, looking at the fact that inference time increases with the number of possible labels. They use \CP on top of a base, simple and fast, text classifier to reduce the number of possible labels for the final, more complex, language model. They experiment on different classification tasks, testing different choices of non-conformity scores and different base models, exploring the trade-off between efficiency and accuracy in choosing the complexity of the base model. 

\paragraph{Speeding up inference.}
To obtain the lightest possible model while preserving performance, \citet{laufer-goldshtein2023efficiently} propose a \CP method to find optimal thresholds to guarantee several risk constraints with adjustable high probability, while optimizing another objective function. They report results on several text classification tasks with different objectives, such as  minimizing prediction cost (searching thresholds on all pruning directions), while controlling accuracy reduction (drop in performance from the full to a lighter model) to a user-chosen degree. Their method builds upon the LTT procedure (\S\ref{subsec:Conformal Risk Control}), with an efficient technique to reduce the number of parameter combinations tested, using Pareto-optimal solutions \citep{pareto}. The results show significant efficiency gains with the proposed risk-controlling guarantees. 

\citet{schuster2022confident} make text generation more efficient by considering decoder early exiting at the token level, while bounding global efficiency. They leverage the LTT procedure to obtain risk-controlling solutions with dynamic allocation of compute per generated token and test their approach on news summarization, text translation and open question answering, showing efficiency gains with the required quality guarantees.

\section{Future Directions}
\label{sec:future}

We outline in this section some promising future research directions and open challenges related to the use of \CP and its many variants in NLP tasks. 

\subsection{\CP for Human-Computer Interaction} 

Some tasks in NLP, such as recommendation and predictive writing systems, benefit naturally from prediction sets that can be used to offer suggestions to users. \CP is an opportunity for improving the efficiency and quality of such systems and prediction sets can be used to enhance performance in decision-making with humans in the loop \citep{cresswell2024conformal}. This aspect could be further explored in NLP, as there are numerous scenarios involving human feedback, \textit{e.g.}, interactive MT \citep{green2013efficacy,wang2021putting} or creation of human preference data for LLM alignment \citep{stiennon2020learning,fernandes2023bridging}.

\subsection{\CP for Handling Label Variation} 

The complexity and ambiguity of natural language, as well as the varied human perspectives, make it hard to disentangle model uncertainty from valid, naturally occurring label variation \citep{baan2024interpreting,plank2022problem,baan2022stop}. It is often the case that multiple outputs are correct, particularly in tasks involving high variation in human language production (question answering, summarization, and other generation tasks where several output variants are equivalent) or inherent, plausible disagreement (the ChaosNLI data that demonstrates valid disagreements in textual inference annotations \cite{pavlick-kwiatkowski-2019-inherent}). 
While traditional methods focus on the majority class, or see variation as model uncertainty, \CP yields a more faithful representation of label variation. Besides representing uncertainty, the sets produced by CP provide multiple ``equivalent'' labels, allowing for more interpretable and informed predictions. Further research on such scenarios could provide models that behave better in tasks with high label variation. Moreover, in such cases, \CP can also be used to achieve diverse prediction sets, avoiding redundancy, as suggested by \citet{quach2024conformal}.

\subsection{\CP for Fairness}
The increased use of NLP systems in global daily life and high-risk tasks raises concerns about the fairness of their outputs. Many of these systems have been shown to be \textit{biased} \citep{blodgett2020language}. In tasks such as resume filtering, medical diagnosis assistance, and several others, these biases can be extremely harmful, leading to skewed performance and coverage. \CP can be used to achieve equalized coverage for different population groups \citep{romano2019malice}, thus ``correcting'' biases in model predictions without the need for expensive retraining. The open research problem of finding conditional guarantees \citep{gibbs2023conformal} to obtain pointwise error bounds can also contribute towards fairness in NLP applications.

\subsection{CP for Dealing with Data Limitations} 

Learning and quantifying uncertainty with limited data is challenging, particularly in NLP problems where manual text labeling can be difficult, time-consuming, and expensive. Approaches to leverage limited data, such as {active learning}, make use of uncertainty quantification in order to reduce the need for manual labelling \citep{Settles2009ActiveLL}. In these settings, \CP could be used for reliable uncertainty quantification, \textit{e.g.}, selecting points with larger prediction sets for manual labelling. The predicted sets can also be useful to reduce the possible labels in tasks with high cardinality output spaces, increasing the performance of subsequent predictions.
Another option is to use \CP for data filtering and cleaning to increase the performance of LLMs \citep{data_pruning}, using for example a small reliable set for calibration, in order to identify mislabeled or noisy samples.

\subsection{\CP for Uncertainty-Aware Evaluation} 

\CP is also useful for tackling the current challenge of model evaluation. 
There are some concerns regarding the current way NLP systems are evaluated: \textit{e.g.}, questioning  how confident we can be in evaluations that result from an LLM scoring the output of another one. Evaluating a conformal predictor built on top of a predictor can be a more reliable way to assess model performance. Another useful application of \CP is to compare different uncertainty heuristics and transformations of model outputs by designing distinct non-conformity scores and evaluating the efficiency (\textit{e.g.}, set size, conditional coverage, observed fuzziness) of the resulting predictors (\S\ref{subsec:app_eval}).

\subsection{Open Challenges}

Despite its numerous applications, using \CP in NLP poses challenges, particularly in generation tasks, providing exciting areas for further research.

\paragraph{Token level.} The non-exchangeability of the data tackled by \citet{barber2023conformal}, \citet{ulmer2024nonexchangeable}, and \citet{farinhas2024nonexchangeable} still presents an obstacle since it is not currently easy to: quantify the coverage gap---the bound in Eq.~\ref{eq:non-x-cp} involves computing a total variation distance between unknown distributions, which is hard to estimate; find good strategies for choosing the weights. 
 \paragraph{Sentence level.} The high cardinality of the output space in generation tasks raises a challenge to typical \CP applications. There are open questions on how to sample the possible outputs and on what is 
 the impact of considering a finite set of samples. 

\section{Conclusion}
This paper provides an overview of applications of the conformal prediction framework in NLP tasks, after a brief introduction to that framework and its main variants. We showed how conformal prediction is a promising tool to address the uncertainty quantification challenge in NLP and hope the existing and possible applications presented in this survey will motivate future research on the topic.


\clearpage

\bibliography{references}

\begin{thebibliography}{101}
\expandafter\ifx\csname natexlab\endcsname\relax\def\natexlab#1{#1}\fi

\bibitem[{Angelopoulos and Bates(2023)}]{tutorial_gentle}
Anastasios~N. Angelopoulos and Stephen Bates. 2023.
\newblock \href {https://doi.org/10.1561/2200000101} {Conformal prediction: A
  gentle introduction}.
\newblock \emph{Foundations and Trends in Machine Learning}, 16(4):494–591.

\bibitem[{Angelopoulos et~al.(2022)Angelopoulos, Bates, Candès, Jordan, and
  Lei}]{angelopoulos2022learn}
Anastasios~N. Angelopoulos, Stephen Bates, Emmanuel~J. Candès, Michael~I.
  Jordan, and Lihua Lei. 2022.
\newblock \href {https://arxiv.org/abs/2110.01052} {Learn then test:
  Calibrating predictive algorithms to achieve risk control}.
\newblock \emph{arXiv preprint arXiv:2110.01052}.

\bibitem[{Angelopoulos et~al.(2024)Angelopoulos, Bates, Fisch, Lei, and
  Schuster}]{angelopoulos2024conformal}
Anastasios~N. Angelopoulos, Stephen Bates, Adam Fisch, Lihua Lei, and Tal
  Schuster. 2024.
\newblock \href {https://openreview.net/forum?id=33XGfHLtZg} {Conformal risk
  control}.
\newblock In \emph{The Twelfth International Conference on Learning
  Representations}.

\bibitem[{Angelopoulos et~al.(2023)Angelopoulos, Candes, and
  Tibshirani}]{angelopoulos2023conformalpid}
Anastasios~Nikolas Angelopoulos, Emmanuel Candes, and Ryan Tibshirani. 2023.
\newblock \href {https://openreview.net/forum?id=zPYeYv6YYs} {Conformal {PID}
  control for time series prediction}.
\newblock In \emph{Thirty-seventh Conference on Neural Information Processing
  Systems}.

\bibitem[{Baan et~al.(2022)Baan, Aziz, Plank, and Fernandez}]{baan2022stop}
Joris Baan, Wilker Aziz, Barbara Plank, and Raquel Fernandez. 2022.
\newblock \href {https://doi.org/10.18653/v1/2022.emnlp-main.124} {Stop
  measuring calibration when humans disagree}.
\newblock In \emph{Proceedings of the 2022 Conference on Empirical Methods in
  Natural Language Processing}, pages 1892--1915, Abu Dhabi, United Arab
  Emirates. Association for Computational Linguistics.

\bibitem[{Baan et~al.(2023)Baan, Daheim, Ilia, Ulmer, Li, Fernández, Plank,
  Sennrich, Zerva, and Aziz}]{baan2023uncertainty}
Joris Baan, Nico Daheim, Evgenia Ilia, Dennis Ulmer, Haau-Sing Li, Raquel
  Fernández, Barbara Plank, Rico Sennrich, Chrysoula Zerva, and Wilker Aziz.
  2023.
\newblock \href {https://arxiv.org/abs/2307.15703} {Uncertainty in natural
  language generation: From theory to applications}.
\newblock \emph{arXiv preprint arXiv:2307.15703}.

\bibitem[{Baan et~al.(2024)Baan, Fern{\'a}ndez, Plank, and
  Aziz}]{baan2024interpreting}
Joris Baan, Raquel Fern{\'a}ndez, Barbara Plank, and Wilker Aziz. 2024.
\newblock \href {https://aclanthology.org/2024.eacl-short.24} {Interpreting
  predictive probabilities: Model confidence or human label variation?}
\newblock In \emph{Proceedings of the 18th Conference of the European Chapter
  of the Association for Computational Linguistics (Volume 2: Short Papers)},
  pages 268--277, St. Julian{'}s, Malta. Association for Computational
  Linguistics.

\bibitem[{Barber et~al.(2020)Barber, Candès, Ramdas, and
  Tibshirani}]{conditional_limits}
Rina~Foygel Barber, Emmanuel~J Candès, Aaditya Ramdas, and Ryan~J Tibshirani.
  2020.
\newblock \href {https://doi.org/10.1093/imaiai/iaaa017} {The limits of
  distribution-free conditional predictive inference}.
\newblock \emph{Information and Inference: A Journal of the IMA},
  10(1):455--482.

\bibitem[{Barber et~al.(2021)Barber, Candès, Ramdas, and
  Tibshirani}]{barber2020predictive}
Rina~Foygel Barber, Emmanuel~J. Candès, Aaditya Ramdas, and Ryan~J.
  Tibshirani. 2021.
\newblock \href {https://doi.org/10.1214/20-aos1965} {Predictive inference with
  the jackknife+}.
\newblock \emph{The Annals of Statistics}, 49(1).

\bibitem[{Barber et~al.(2023)Barber, Candès, Ramdas, and
  Tibshirani}]{barber2023conformal}
Rina~Foygel Barber, Emmanuel~J. Candès, Aaditya Ramdas, and Ryan~J.
  Tibshirani. 2023.
\newblock \href {https://doi.org/10.1214/23-aos2276} {Conformal prediction
  beyond exchangeability}.
\newblock \emph{The Annals of Statistics}, 51(2).

\bibitem[{Bates et~al.(2021)Bates, Angelopoulos, Lei, Malik, and
  Jordan}]{bates2021rcps}
Stephen Bates, Anastasios Angelopoulos, Lihua Lei, Jitendra Malik, and Michael
  Jordan. 2021.
\newblock \href {https://doi.org/10.1145/3478535} {Distribution-free,
  risk-controlling prediction sets}.
\newblock \emph{J. ACM}, 68(6).

\bibitem[{Blodgett et~al.(2020)Blodgett, Barocas, Daum{\'e}~III, and
  Wallach}]{blodgett2020language}
Su~Lin Blodgett, Solon Barocas, Hal Daum{\'e}~III, and Hanna Wallach. 2020.
\newblock \href {https://doi.org/10.18653/v1/2020.acl-main.485} {Language
  (technology) is power: A critical survey of {``}bias{''} in {NLP}}.
\newblock In \emph{Proceedings of the 58th Annual Meeting of the Association
  for Computational Linguistics}, pages 5454--5476, Online. Association for
  Computational Linguistics.

\bibitem[{Chernozhukov et~al.(2018)Chernozhukov, W\"{u}thrich, and
  Yinchu}]{chernozhukov2018exact}
Victor Chernozhukov, Kaspar W\"{u}thrich, and Zhu Yinchu. 2018.
\newblock \href {https://proceedings.mlr.press/v75/chernozhukov18a.html} {Exact
  and robust conformal inference methods for predictive machine learning with
  dependent data}.
\newblock In \emph{Proceedings of the 31st Conference On Learning Theory},
  volume~75 of \emph{Proceedings of Machine Learning Research}, pages 732--749.
  PMLR.

\bibitem[{Choubey et~al.(2022)Choubey, Bai, Wu, Liu, and
  Rajani}]{choubey-etal-2022-conformal}
Prafulla~Kumar Choubey, Yu~Bai, Chien-Sheng Wu, Wenhao Liu, and Nazneen Rajani.
  2022.
\newblock \href {https://doi.org/10.18653/v1/2022.emnlp-main.196} {Conformal
  predictor for improving zero-shot text classification efficiency}.
\newblock In \emph{Proceedings of the 2022 Conference on Empirical Methods in
  Natural Language Processing}, pages 3027--3034, Abu Dhabi, United Arab
  Emirates. Association for Computational Linguistics.

\bibitem[{Cresswell et~al.(2024)Cresswell, Sui, Kumar, and
  Vouitsis}]{cresswell2024conformal}
Jesse~C. Cresswell, Yi~Sui, Bhargava Kumar, and Noël Vouitsis. 2024.
\newblock \href {https://arxiv.org/abs/2401.13744} {Conformal prediction sets
  improve human decision making}.
\newblock \emph{arXiv preprint arXiv:2401.13744}.

\bibitem[{Deb and Kalyanmoy(2001)}]{pareto}
Kalyanmoy Deb and Deb Kalyanmoy. 2001.
\newblock \emph{Multi-Objective Optimization Using Evolutionary Algorithms}.
\newblock John Wiley \& Sons, Inc., USA.

\bibitem[{Deng et~al.(2020)Deng, Li, Han, Shi, and Xie}]{survey_acceleration}
Lei Deng, Guoqi Li, Song Han, Luping Shi, and Yuan Xie. 2020.
\newblock \href {https://doi.org/10.1109/JPROC.2020.2976475} {Model compression
  and hardware acceleration for neural networks: A comprehensive survey}.
\newblock \emph{Proceedings of the IEEE}, 108(4):485--532.

\bibitem[{Desai and Durrett(2020)}]{desai-durrett-2020-calibration}
Shrey Desai and Greg Durrett. 2020.
\newblock \href {https://doi.org/10.18653/v1/2020.emnlp-main.21} {Calibration
  of pre-trained transformers}.
\newblock In \emph{Proceedings of the 2020 Conference on Empirical Methods in
  Natural Language Processing (EMNLP)}, pages 295--302, Online. Association for
  Computational Linguistics.

\bibitem[{Devlin et~al.(2019)Devlin, Chang, Lee, and
  Toutanova}]{devlin2019bert}
Jacob Devlin, Ming-Wei Chang, Kenton Lee, and Kristina Toutanova. 2019.
\newblock \href {https://doi.org/10.18653/v1/N19-1423} {{BERT}: Pre-training of
  deep bidirectional transformers for language understanding}.
\newblock In \emph{Proceedings of the 2019 Conference of the North {A}merican
  Chapter of the Association for Computational Linguistics: Human Language
  Technologies, Volume 1 (Long and Short Papers)}, pages 4171--4186,
  Minneapolis, Minnesota. Association for Computational Linguistics.

\bibitem[{Dey et~al.(2022)Dey, Ding, Ferrell, Kapper, Lovig, Planchon, and
  Williams}]{dey2021conformal}
Neil Dey, Jing Ding, Jack Ferrell, Carolina Kapper, Maxwell Lovig, Emiliano
  Planchon, and Jonathan~P. Williams. 2022.
\newblock \href {https://doi.org/10.51387/22-NEJSDS8} {Conformal prediction for
  text infilling and part-of-speech prediction}.
\newblock \emph{The New England Journal of Statistics in Data Science},
  1(1):69--83.

\bibitem[{Dutta et~al.(2023)Dutta, Wei, van~der Laan, and
  Alaa}]{dutta2023estimating}
Shiladitya Dutta, Hongbo Wei, Lars van~der Laan, and Ahmed Alaa. 2023.
\newblock \href {https://openreview.net/forum?id=f7mLYe95m4} {Estimating
  uncertainty in multimodal foundation models using public internet data}.
\newblock In \emph{R0-FoMo:Robustness of Few-shot and Zero-shot Learning in
  Large Foundation Models}.

\bibitem[{Ernez et~al.(2023)Ernez, Arnold, Galametz, Kobus, and
  Ould-Amer}]{ernez23applying}
Fares Ernez, Alexandre Arnold, Audrey Galametz, Catherine Kobus, and Nawal
  Ould-Amer. 2023.
\newblock \href {https://proceedings.mlr.press/v204/ernez23a.html} {Applying
  the conformal prediction paradigm for the uncertainty quantification of an
  end-to-end automatic speech recognition model (wav2vec 2.0)}.
\newblock In \emph{Proceedings of the Twelfth Symposium on Conformal and
  Probabilistic Prediction with Applications}, volume 204 of \emph{Proceedings
  of Machine Learning Research}, pages 16--35. PMLR.

\bibitem[{Farinhas et~al.(2024)Farinhas, Zerva, Ulmer, and
  Martins}]{farinhas2024nonexchangeable}
António Farinhas, Chrysoula Zerva, Dennis Ulmer, and André F.~T. Martins.
  2024.
\newblock \href {https://openreview.net/forum?id=j511LaqEeP} {Non-exchangeable
  conformal risk control}.
\newblock In \emph{The Twelfth International Conference on Learning
  Representations}.

\bibitem[{Feldman et~al.(2021)Feldman, Bates, and
  Romano}]{feldman2021improving}
Shai Feldman, Stephen Bates, and Yaniv Romano. 2021.
\newblock \href {https://openreview.net/forum?id=pTe-8qCdDqy} {Improving
  conditional coverage via orthogonal quantile regression}.
\newblock In \emph{Advances in Neural Information Processing Systems}.

\bibitem[{Fernandes et~al.(2023)Fernandes, Madaan, Liu, Farinhas, Martins,
  Bertsch, de~Souza, Zhou, Wu, Neubig et~al.}]{fernandes2023bridging}
Patrick Fernandes, Aman Madaan, Emmy Liu, Ant{\'o}nio Farinhas, Pedro~Henrique
  Martins, Amanda Bertsch, Jos{\'e}~GC de~Souza, Shuyan Zhou, Tongshuang Wu,
  Graham Neubig, et~al. 2023.
\newblock Bridging the gap: A survey on integrating (human) feedback for
  natural language generation.
\newblock \emph{Transactions of the Association for Computational Linguistics},
  11:1643--1668.

\bibitem[{Fisch et~al.(2021{\natexlab{a}})Fisch, Schuster, Jaakkola, and
  Barzilay}]{fisch2021fewshot}
Adam Fisch, Tal Schuster, Tommi Jaakkola, and Dr.Regina Barzilay.
  2021{\natexlab{a}}.
\newblock \href {https://proceedings.mlr.press/v139/fisch21a.html} {Few-shot
  conformal prediction with auxiliary tasks}.
\newblock In \emph{Proceedings of the 38th International Conference on Machine
  Learning}, volume 139 of \emph{Proceedings of Machine Learning Research},
  pages 3329--3339. PMLR.

\bibitem[{Fisch et~al.(2022)Fisch, Schuster, Jaakkola, and
  Barzilay}]{fisch2022fp}
Adam Fisch, Tal Schuster, Tommi Jaakkola, and Dr.Regina Barzilay. 2022.
\newblock \href {https://proceedings.mlr.press/v162/fisch22a.html} {Conformal
  prediction sets with limited false positives}.
\newblock In \emph{Proceedings of the 39th International Conference on Machine
  Learning}, volume 162 of \emph{Proceedings of Machine Learning Research},
  pages 6514--6532. PMLR.

\bibitem[{Fisch et~al.(2021{\natexlab{b}})Fisch, Schuster, Jaakkola, and
  Barzilay}]{fisch2021efficient}
Adam Fisch, Tal Schuster, Tommi~S. Jaakkola, and Regina Barzilay.
  2021{\natexlab{b}}.
\newblock \href {https://openreview.net/forum?id=tnSo6VRLmT} {Efficient
  conformal prediction via cascaded inference with expanded admission}.
\newblock In \emph{International Conference on Learning Representations}.

\bibitem[{Gallegos et~al.(2024)Gallegos, Rossi, Barrow, Tanjim, Kim,
  Dernoncourt, Yu, Zhang, and Ahmed}]{gallegos2024bias}
Isabel~O. Gallegos, Ryan~A. Rossi, Joe Barrow, Md~Mehrab Tanjim, Sungchul Kim,
  Franck Dernoncourt, Tong Yu, Ruiyi Zhang, and Nesreen~K. Ahmed. 2024.
\newblock \href {https://arxiv.org/abs/2309.00770} {Bias and fairness in large
  language models: A survey}.
\newblock \emph{arXiv preprint arXiv:2309.00770}.

\bibitem[{Gibbs and Candes(2021)}]{gibbs2021adaptive}
Isaac Gibbs and Emmanuel Candes. 2021.
\newblock \href
  {https://proceedings.neurips.cc/paper_files/paper/2021/file/0d441de75945e5acbc865406fc9a2559-Paper.pdf}
  {Adaptive conformal inference under distribution shift}.
\newblock In \emph{Advances in Neural Information Processing Systems},
  volume~34, pages 1660--1672. Curran Associates, Inc.

\bibitem[{Gibbs et~al.(2023)Gibbs, Cherian, and Candès}]{gibbs2023conformal}
Isaac Gibbs, John~J. Cherian, and Emmanuel~J. Candès. 2023.
\newblock \href {https://arxiv.org/abs/2305.12616} {Conformal prediction with
  conditional guarantees}.
\newblock \emph{arXiv preprint arXiv:2305.12616}.

\bibitem[{Giovannotti(2022)}]{giovannotti2022calibration}
Patrizio Giovannotti. 2022.
\newblock \href {https://proceedings.mlr.press/v179/giovannotti22a.html}
  {Calibration of natural language understanding models with venn–abers
  predictors}.
\newblock In \emph{Proceedings of the Eleventh Symposium on Conformal and
  Probabilistic Prediction with Applications}, volume 179 of \emph{Proceedings
  of Machine Learning Research}, pages 55--71. PMLR.

\bibitem[{Giovannotti(2023)}]{giovannotti2023evaluating}
Patrizio Giovannotti. 2023.
\newblock \href {https://proceedings.mlr.press/v204/giovannotti23a.html}
  {Evaluating machine translation quality with conformal predictive
  distributions}.
\newblock In \emph{Proceedings of the Twelfth Symposium on Conformal and
  Probabilistic Prediction with Applications}, volume 204 of \emph{Proceedings
  of Machine Learning Research}, pages 413--429. PMLR.

\bibitem[{Giovannotti and Gammerman(2021)}]{giovannotti2021paraphrase}
Patrizio Giovannotti and Alex Gammerman. 2021.
\newblock \href {https://proceedings.mlr.press/v152/giovannotti21a.html}
  {Transformer-based conformal predictors for paraphrase detection}.
\newblock In \emph{Proceedings of the Tenth Symposium on Conformal and
  Probabilistic Prediction and Applications}, volume 152 of \emph{Proceedings
  of Machine Learning Research}, pages 243--265. PMLR.

\bibitem[{Glushkova et~al.(2021)Glushkova, Zerva, Rei, and
  Martins}]{glushkova-etal-2021-uncertainty-aware}
Taisiya Glushkova, Chrysoula Zerva, Ricardo Rei, and Andr{\'e} F.~T. Martins.
  2021.
\newblock \href {https://doi.org/10.18653/v1/2021.findings-emnlp.330}
  {Uncertainty-aware machine translation evaluation}.
\newblock In \emph{Findings of the Association for Computational Linguistics:
  EMNLP 2021}, pages 3920--3938, Punta Cana, Dominican Republic. Association
  for Computational Linguistics.

\bibitem[{Golchin and Surdeanu(2024)}]{contamination_2}
Shahriar Golchin and Mihai Surdeanu. 2024.
\newblock \href {https://openreview.net/forum?id=2Rwq6c3tvr} {Time travel in
  {LLM}s: Tracing data contamination in large language models}.
\newblock In \emph{The Twelfth International Conference on Learning
  Representations}.

\bibitem[{Green et~al.(2013)Green, Heer, and Manning}]{green2013efficacy}
Spence Green, Jeffrey Heer, and Christopher~D Manning. 2013.
\newblock The efficacy of human post-editing for language translation.
\newblock In \emph{Proceedings of the SIGCHI conference on human factors in
  computing systems}, pages 439--448.

\bibitem[{Guerreiro et~al.(2023)Guerreiro, Alves, Waldendorf, Haddow, Birch,
  Colombo, and Martins}]{guerreiro2023hallucinations}
Nuno~M. Guerreiro, Duarte~M. Alves, Jonas Waldendorf, Barry Haddow, Alexandra
  Birch, Pierre Colombo, and André F.~T. Martins. 2023.
\newblock \href {https://doi.org/10.1162/tacl_a_00615} {{Hallucinations in
  Large Multilingual Translation Models}}.
\newblock \emph{Transactions of the Association for Computational Linguistics},
  11:1500--1517.

\bibitem[{Hanu and {Unitary team}(2020)}]{Detoxify}
Laura Hanu and {Unitary team}. 2020.
\newblock \href {https://github.com/unitaryai/detoxify} {Detoxify}.

\bibitem[{He et~al.(2020)He, Zhang, Lei, Chen, Chen, Alhamadani, Xiao, and
  Lu}]{he-etal-2020-towards}
Jianfeng He, Xuchao Zhang, Shuo Lei, Zhiqian Chen, Fanglan Chen, Abdulaziz
  Alhamadani, Bei Xiao, and ChangTien Lu. 2020.
\newblock \href {https://doi.org/10.18653/v1/2020.emnlp-main.671} {Towards more
  accurate uncertainty estimation in text classification}.
\newblock In \emph{Proceedings of the 2020 Conference on Empirical Methods in
  Natural Language Processing (EMNLP)}, pages 8362--8372, Online. Association
  for Computational Linguistics.

\bibitem[{Hechtlinger et~al.(2019)Hechtlinger, Póczos, and
  Wasserman}]{hechtlinger2019cautious}
Yotam Hechtlinger, Barnabás Póczos, and Larry Wasserman. 2019.
\newblock \href {https://arxiv.org/abs/1805.09460} {Cautious deep learning}.
\newblock \emph{arXiv preprint arXiv:1805.09460}.

\bibitem[{Holtzman et~al.(2020)Holtzman, Buys, Du, Forbes, and
  Choi}]{Holtzman2020The}
Ari Holtzman, Jan Buys, Li~Du, Maxwell Forbes, and Yejin Choi. 2020.
\newblock \href {https://openreview.net/forum?id=rygGQyrFvH} {The curious case
  of neural text degeneration}.
\newblock In \emph{International Conference on Learning Representations}.

\bibitem[{Hu et~al.(2023)Hu, Zhang, Zhao, Huang, and Wu}]{hu2023uncertainty}
Mengting Hu, Zhen Zhang, Shiwan Zhao, Minlie Huang, and Bingzhe Wu. 2023.
\newblock \href {https://arxiv.org/abs/2306.04459} {Uncertainty in natural
  language processing: Sources, quantification, and applications}.
\newblock \emph{arXiv preprint arXiv:2306.04459}.

\bibitem[{Huang et~al.(2023)Huang, Yu, Ma, Zhong, Feng, Wang, Chen, Peng, Feng,
  Qin, and Liu}]{huang2023survey}
Lei Huang, Weijiang Yu, Weitao Ma, Weihong Zhong, Zhangyin Feng, Haotian Wang,
  Qianglong Chen, Weihua Peng, Xiaocheng Feng, Bing Qin, and Ting Liu. 2023.
\newblock \href {https://arxiv.org/abs/2311.05232} {A survey on hallucination
  in large language models: Principles, taxonomy, challenges, and open
  questions}.
\newblock \emph{arXiv preprint arXiv:2311.05232}.

\bibitem[{Ji et~al.(2023)Ji, Lee, Frieske, Yu, Su, Xu, Ishii, Bang, Madotto,
  and Fung}]{ji2023survey}
Ziwei Ji, Nayeon Lee, Rita Frieske, Tiezheng Yu, Dan Su, Yan Xu, Etsuko Ishii,
  Ye~Jin Bang, Andrea Madotto, and Pascale Fung. 2023.
\newblock \href {https://doi.org/10.1145/3571730} {Survey of hallucination in
  natural language generation}.
\newblock \emph{ACM Comput. Surv.}, 55(12).

\bibitem[{Jin and Ren(2024)}]{jin2024confidence}
Ying Jin and Zhimei Ren. 2024.
\newblock \href {https://arxiv.org/abs/2403.03868} {Confidence on the focal:
  Conformal prediction with selection-conditional coverage}.
\newblock \emph{arXiv preprint arXiv:2403.03868}.

\bibitem[{Johansson et~al.(2023)Johansson, Löfström, and
  Sönströd}]{johansson2023wellcalibrated}
Ulf Johansson, Tuwe Löfström, and Cecilia Sönströd. 2023.
\newblock \href {https://arxiv.org/abs/2306.06642} {Well-calibrated
  probabilistic predictive maintenance using venn-abers}.
\newblock \emph{arXiv preprint arXiv:2306.06642}.

\bibitem[{Kulal et~al.(2019)Kulal, Pasupat, Chandra, Lee, Padon, Aiken, and
  Liang}]{kulal2019spoc}
Sumith Kulal, Panupong Pasupat, Kartik Chandra, Mina Lee, Oded Padon, Alex
  Aiken, and Percy~S Liang. 2019.
\newblock \href
  {https://proceedings.neurips.cc/paper_files/paper/2019/file/7298332f04ac004a0ca44cc69ecf6f6b-Paper.pdf}
  {Spoc: Search-based pseudocode to code}.
\newblock In \emph{Advances in Neural Information Processing Systems},
  volume~32. Curran Associates, Inc.

\bibitem[{Kuleshov et~al.(2018)Kuleshov, Fenner, and
  Ermon}]{pmlr-v80-kuleshov18a}
Volodymyr Kuleshov, Nathan Fenner, and Stefano Ermon. 2018.
\newblock \href {https://proceedings.mlr.press/v80/kuleshov18a.html} {Accurate
  uncertainties for deep learning using calibrated regression}.
\newblock In \emph{Proceedings of the 35th International Conference on Machine
  Learning}, volume~80 of \emph{Proceedings of Machine Learning Research},
  pages 2796--2804. PMLR.

\bibitem[{Kumar et~al.(2023)Kumar, Lu, Gupta, Palepu, Bellamy, Raskar, and
  Beam}]{kumar2023conformal}
Bhawesh Kumar, Charlie Lu, Gauri Gupta, Anil Palepu, David Bellamy, Ramesh
  Raskar, and Andrew Beam. 2023.
\newblock \href {https://arxiv.org/abs/2305.18404} {Conformal prediction with
  large language models for multi-choice question answering}.
\newblock \emph{arXiv preprint arXiv:2305.18404}.

\bibitem[{Lambrou et~al.(2014)Lambrou, Nouretdinov, and
  Papadopoulos}]{lambrou2014ivap}
Antonis Lambrou, Ilia Nouretdinov, and Harris Papadopoulos. 2014.
\newblock \href {https://doi.org/10.1007/s10472-014-9420-z} {Inductive venn
  prediction}.
\newblock \emph{Annals of Mathematics and Artificial Intelligence}, 74.

\bibitem[{Laufer-Goldshtein et~al.(2023)Laufer-Goldshtein, Fisch, Barzilay, and
  Jaakkola}]{laufer-goldshtein2023efficiently}
Bracha Laufer-Goldshtein, Adam Fisch, Regina Barzilay, and Tommi~S. Jaakkola.
  2023.
\newblock \href {https://openreview.net/forum?id=cyg2YXn_BqF} {Efficiently
  controlling multiple risks with pareto testing}.
\newblock In \emph{The Eleventh International Conference on Learning
  Representations}.

\bibitem[{Leake(2012)}]{Leake2012}
David~B. Leake. 2012.
\newblock \href {https://doi.org/10.1007/978-1-4419-1428-6_1802}
  {\emph{Introspective Learning and Reasoning}}. Springer US, Boston, MA.

\bibitem[{Lei et~al.(2018)Lei, G'Sell, Rinaldo, Tibshirani, and
  Wasserman}]{lei2017distributionfree}
Jing Lei, Max G'Sell, Alessandro Rinaldo, Ryan~J. Tibshirani, and Larry
  Wasserman. 2018.
\newblock \href {https://doi.org/10.1080/01621459.2017.1307116}
  {Distribution-free predictive inference for regression}.
\newblock \emph{Journal of the American Statistical Association},
  113(523):1094--1111.

\bibitem[{Liang et~al.(2024)Liang, Zhang, and Fisac}]{liang2024introspective}
Kaiqu Liang, Zixu Zhang, and Jaime~Fernández Fisac. 2024.
\newblock \href {https://arxiv.org/abs/2402.06529} {Introspective planning:
  Guiding language-enabled agents to refine their own uncertainty}.
\newblock \emph{arXiv preprint arXiv:2402.06529}.

\bibitem[{Lin(2004)}]{lin-2004-rouge}
Chin-Yew Lin. 2004.
\newblock \href {https://aclanthology.org/W04-1013} {{ROUGE}: A package for
  automatic evaluation of summaries}.
\newblock In \emph{Text Summarization Branches Out}, pages 74--81, Barcelona,
  Spain. Association for Computational Linguistics.

\bibitem[{Liu et~al.(2019)Liu, Ott, Goyal, Du, Joshi, Chen, Levy, Lewis,
  Zettlemoyer, and Stoyanov}]{liu2019earlyexit}
Yinhan Liu, Myle Ott, Naman Goyal, Jingfei Du, Mandar Joshi, Danqi Chen, Omer
  Levy, Mike Lewis, Luke Zettlemoyer, and Veselin Stoyanov. 2019.
\newblock \href {https://arxiv.org/abs/1907.11692} {Roberta: A robustly
  optimized bert pretraining approach}.
\newblock \emph{arXiv preprint arXiv:1907.11692}.

\bibitem[{Maltoudoglou et~al.(2022)Maltoudoglou, Paisios, Lenc, Martínek,
  Král, and Papadopoulos}]{maltoudoglou2022multilabel}
Lysimachos Maltoudoglou, Andreas Paisios, Ladislav Lenc, Jiří Martínek,
  Pavel Král, and Harris Papadopoulos. 2022.
\newblock \href {https://doi.org/10.1016/j.patcog.2021.108271} {Well-calibrated
  confidence measures for multi-label text classification with a large number
  of labels}.
\newblock \emph{Pattern Recognition}, 122:108271.

\bibitem[{Maltoudoglou et~al.(2020)Maltoudoglou, Paisios, and
  Papadopoulos}]{paisios2020bert}
Lysimachos Maltoudoglou, Andreas Paisios, and Harris Papadopoulos. 2020.
\newblock \href {https://proceedings.mlr.press/v128/maltoudoglou20a.html}
  {Bert-based conformal predictor for sentiment analysis}.
\newblock In \emph{Proceedings of the Ninth Symposium on Conformal and
  Probabilistic Prediction and Applications}, volume 128 of \emph{Proceedings
  of Machine Learning Research}, pages 269--284. PMLR.

\bibitem[{Manokhin(2017)}]{multi_class_ivap}
Valery Manokhin. 2017.
\newblock \href {https://proceedings.mlr.press/v60/manokhin17a.html}
  {Multi-class probabilistic classification using inductive and cross
  {V}enn–{A}bers predictors}.
\newblock In \emph{Proceedings of the Sixth Workshop on Conformal and
  Probabilistic Prediction and Applications}, volume~60 of \emph{Proceedings of
  Machine Learning Research}, pages 228--240. PMLR.

\bibitem[{Marion et~al.(2023)Marion, Üstün, Pozzobon, Wang, Fadaee, and
  Hooker}]{data_pruning}
Max Marion, Ahmet Üstün, Luiza Pozzobon, Alex Wang, Marzieh Fadaee, and Sara
  Hooker. 2023.
\newblock \href {https://openreview.net/forum?id=XUIYn3jo5T} {When less is
  more: Investigating data pruning for pretraining {LLM}s at scale}.
\newblock In \emph{NeurIPS Workshop on Attributing Model Behavior at Scale}.

\bibitem[{Messoudi et~al.(2020)Messoudi, Rousseau, and
  Destercke}]{messoudi2020deep}
Soundouss Messoudi, Sylvain Rousseau, and Sebastien Destercke. 2020.
\newblock \href {https://doi.org/10.1007/978-3-030-50146-4_39} {\emph{Deep
  Conformal Prediction for Robust Models}}.

\bibitem[{Min et~al.(2023)Min, Ross, Sulem, Veyseh, Nguyen, Sainz, Agirre,
  Heintz, and Roth}]{min2021recent}
Bonan Min, Hayley Ross, Elior Sulem, Amir Pouran~Ben Veyseh, Thien~Huu Nguyen,
  Oscar Sainz, Eneko Agirre, Ilana Heintz, and Dan Roth. 2023.
\newblock \href {https://doi.org/10.1145/3605943} {Recent advances in natural
  language processing via large pre-trained language models: A survey}.
\newblock \emph{ACM Comput. Surv.}, 56(2).

\bibitem[{Norinder and Norinder(2022)}]{ulf2022sa}
Ulf Norinder and Petra Norinder. 2022.
\newblock \href {https://doi.org/10.1080/23270012.2022.2031324} {Predicting
  amazon customer reviews with deep confidence using deep learning and
  conformal prediction}.
\newblock \emph{Journal of Management Analytics}, 9(1):1--16.

\bibitem[{Paisios et~al.(2019)Paisios, Lenc, Mart\'{\i}nek, Kr\'al, and
  Papadopoulos}]{paisios2019multilabel}
Andreas Paisios, Ladislav Lenc, Ji\v{r}\'{\i} Mart\'{\i}nek, Pavel Kr\'al, and
  Harris Papadopoulos. 2019.
\newblock \href {https://proceedings.mlr.press/v105/paisios19a.html} {A deep
  neural network conformal predictor for multi-label text classification}.
\newblock In \emph{Proceedings of the Eighth Symposium on Conformal and
  Probabilistic Prediction and Applications}, volume 105 of \emph{Proceedings
  of Machine Learning Research}, pages 228--245. PMLR.

\bibitem[{Pavlick and Kwiatkowski(2019)}]{pavlick-kwiatkowski-2019-inherent}
Ellie Pavlick and Tom Kwiatkowski. 2019.
\newblock \href {https://doi.org/10.1162/tacl_a_00293} {Inherent disagreements
  in human textual inferences}.
\newblock \emph{Transactions of the Association for Computational Linguistics},
  7:677--694.

\bibitem[{Plank(2022)}]{plank2022problem}
Barbara Plank. 2022.
\newblock The “problem” of human label variation: On ground truth in data,
  modeling and evaluation.
\newblock In \emph{Proceedings of the 2022 Conference on Empirical Methods in
  Natural Language Processing}, pages 10671--10682.

\bibitem[{Podkopaev and Ramdas(2021)}]{podkopaev2021label}
Aleksandr Podkopaev and Aaditya Ramdas. 2021.
\newblock \href {https://proceedings.mlr.press/v161/podkopaev21a.html}
  {Distribution-free uncertainty quantification for classification under label
  shift}.
\newblock In \emph{Proceedings of the Thirty-Seventh Conference on Uncertainty
  in Artificial Intelligence}, volume 161 of \emph{Proceedings of Machine
  Learning Research}, pages 844--853. PMLR.

\bibitem[{Quach et~al.(2024)Quach, Fisch, Schuster, Yala, Sohn, Jaakkola, and
  Barzilay}]{quach2024conformal}
Victor Quach, Adam Fisch, Tal Schuster, Adam Yala, Jae~Ho Sohn, Tommi~S.
  Jaakkola, and Regina Barzilay. 2024.
\newblock \href {https://openreview.net/forum?id=pzUhfQ74c5} {Conformal
  language modeling}.
\newblock In \emph{The Twelfth International Conference on Learning
  Representations}.

\bibitem[{Radford et~al.(2021)Radford, Kim, Hallacy, Ramesh, Goh, Agarwal,
  Sastry, Askell, Mishkin, Clark, Krueger, and Sutskever}]{radford2021learning}
Alec Radford, Jong~Wook Kim, Chris Hallacy, Aditya Ramesh, Gabriel Goh,
  Sandhini Agarwal, Girish Sastry, Amanda Askell, Pamela Mishkin, Jack Clark,
  Gretchen Krueger, and Ilya Sutskever. 2021.
\newblock \href {https://proceedings.mlr.press/v139/radford21a.html} {Learning
  transferable visual models from natural language supervision}.
\newblock In \emph{Proceedings of the 38th International Conference on Machine
  Learning}, volume 139 of \emph{Proceedings of Machine Learning Research},
  pages 8748--8763. PMLR.

\bibitem[{Ravfogel et~al.(2023)Ravfogel, Goldberg, and
  Goldberger}]{ravfogel-etal-2023-conformal}
Shauli Ravfogel, Yoav Goldberg, and Jacob Goldberger. 2023.
\newblock \href {https://doi.org/10.18653/v1/2023.findings-acl.3} {Conformal
  nucleus sampling}.
\newblock In \emph{Findings of the Association for Computational Linguistics:
  ACL 2023}, pages 27--34, Toronto, Canada. Association for Computational
  Linguistics.

\bibitem[{Ren et~al.(2023)Ren, Dixit, Bodrova, Singh, Tu, Brown, Xu, Takayama,
  Xia, Varley, Xu, Sadigh, Zeng, and Majumdar}]{ren2023robots}
Allen~Z. Ren, Anushri Dixit, Alexandra Bodrova, Sumeet Singh, Stephen Tu, Noah
  Brown, Peng Xu, Leila Takayama, Fei Xia, Jake Varley, Zhenjia Xu, Dorsa
  Sadigh, Andy Zeng, and Anirudha Majumdar. 2023.
\newblock \href {https://openreview.net/forum?id=4ZK8ODNyFXx} {Robots that ask
  for help: Uncertainty alignment for large language model planners}.
\newblock In \emph{7th Annual Conference on Robot Learning}.

\bibitem[{Romano et~al.(2020)Romano, Barber, Sabatti, and Cand{\`
  e}s}]{romano2019malice}
Yaniv Romano, Rina~Foygel Barber, Chiara Sabatti, and Emmanuel Cand{\` e}s.
  2020.
\newblock \href {https://hdsr.mitpress.mit.edu/pub/qedrwcz3} {With {Malice}
  {Toward} {None}: Assessing {Uncertainty} via {Equalized} {Coverage}}.
\newblock \emph{Harvard Data Science Review}, 2(2).

\bibitem[{Sainz et~al.(2023)Sainz, Campos, Garc{\'\i}a-Ferrero, Etxaniz,
  de~Lacalle, and Agirre}]{contamination_1}
Oscar Sainz, Jon Campos, Iker Garc{\'\i}a-Ferrero, Julen Etxaniz, Oier~Lopez
  de~Lacalle, and Eneko Agirre. 2023.
\newblock \href {https://doi.org/10.18653/v1/2023.findings-emnlp.722} {{NLP}
  evaluation in trouble: On the need to measure {LLM} data contamination for
  each benchmark}.
\newblock In \emph{Findings of the Association for Computational Linguistics:
  EMNLP 2023}, pages 10776--10787, Singapore. Association for Computational
  Linguistics.

\bibitem[{Schuster et~al.(2022)Schuster, Fisch, Gupta, Dehghani, Bahri, Tran,
  Tay, and Metzler}]{schuster2022confident}
Tal Schuster, Adam Fisch, Jai Gupta, Mostafa Dehghani, Dara Bahri, Vinh~Q.
  Tran, Yi~Tay, and Donald Metzler. 2022.
\newblock \href {https://openreview.net/forum?id=uLYc4L3C81A} {Confident
  adaptive language modeling}.
\newblock In \emph{Advances in Neural Information Processing Systems}.

\bibitem[{Schuster et~al.(2021)Schuster, Fisch, Jaakkola, and
  Barzilay}]{schuster2021consistent}
Tal Schuster, Adam Fisch, Tommi Jaakkola, and Regina Barzilay. 2021.
\newblock \href {https://arxiv.org/abs/2104.08803} {Consistent accelerated
  inference via confident adaptive transformers}.
\newblock \emph{arXiv preprint arXiv:2104.08803}.

\bibitem[{Schwartz et~al.(2020)Schwartz, Stanovsky, Swayamdipta, Dodge, and
  Smith}]{schwartz2020earlyexit}
Roy Schwartz, Gabriel Stanovsky, Swabha Swayamdipta, Jesse Dodge, and Noah~A.
  Smith. 2020.
\newblock \href {https://doi.org/10.18653/v1/2020.acl-main.593} {The right tool
  for the job: Matching model and instance complexities}.
\newblock In \emph{Proceedings of the 58th Annual Meeting of the Association
  for Computational Linguistics}, pages 6640--6651, Online. Association for
  Computational Linguistics.

\bibitem[{Settles(2009)}]{Settles2009ActiveLL}
Burr Settles. 2009.
\newblock \href {https://api.semanticscholar.org/CorpusID:324600} {Active
  learning literature survey}.

\bibitem[{Shafer and Vovk(2008)}]{tutorial_vovk}
Glenn Shafer and Vladimir Vovk. 2008.
\newblock A tutorial on conformal prediction.
\newblock \emph{J. Mach. Learn. Res.}, 9:371–421.

\bibitem[{Stiennon et~al.(2020)Stiennon, Ouyang, Wu, Ziegler, Lowe, Voss,
  Radford, Amodei, and Christiano}]{stiennon2020learning}
Nisan Stiennon, Long Ouyang, Jeffrey Wu, Daniel Ziegler, Ryan Lowe, Chelsea
  Voss, Alec Radford, Dario Amodei, and Paul~F Christiano. 2020.
\newblock Learning to summarize with human feedback.
\newblock \emph{Advances in Neural Information Processing Systems},
  33:3008--3021.

\bibitem[{Tibshirani et~al.(2019)Tibshirani, Foygel~Barber, Candes, and
  Ramdas}]{tibshirani2019covariate}
Ryan~J Tibshirani, Rina Foygel~Barber, Emmanuel Candes, and Aaditya Ramdas.
  2019.
\newblock \href
  {https://proceedings.neurips.cc/paper_files/paper/2019/file/8fb21ee7a2207526da55a679f0332de2-Paper.pdf}
  {Conformal prediction under covariate shift}.
\newblock In \emph{Advances in Neural Information Processing Systems},
  volume~32. Curran Associates, Inc.

\bibitem[{Tsoumakas et~al.(2010)Tsoumakas, Katakis, and
  Vlahavas}]{Tsoumakas2010}
Grigorios Tsoumakas, Ioannis Katakis, and Ioannis Vlahavas. 2010.
\newblock \href {https://doi.org/10.1007/978-0-387-09823-4_34} {\emph{Mining
  Multi-label Data}}. Springer US, Boston, MA.

\bibitem[{Ulmer et~al.(2024)Ulmer, Zerva, and
  Martins}]{ulmer2024nonexchangeable}
Dennis Ulmer, Chrysoula Zerva, and Andre Martins. 2024.
\newblock \href {https://aclanthology.org/2024.findings-eacl.129}
  {Non-exchangeable conformal language generation with nearest neighbors}.
\newblock In \emph{Findings of the Association for Computational Linguistics:
  EACL 2024}, pages 1909--1929, St. Julian{'}s, Malta. Association for
  Computational Linguistics.

\bibitem[{Vasudevan et~al.(2019)Vasudevan, Sethy, and Ghias}]{uncalibrated}
Vishal~Thanvantri Vasudevan, Abhinav Sethy, and Alireza~Roshan Ghias. 2019.
\newblock \href {https://doi.org/10.1109/ICASSP.2019.8683359} {Towards better
  confidence estimation for neural models}.
\newblock In \emph{ICASSP 2019 - 2019 IEEE International Conference on
  Acoustics, Speech and Signal Processing (ICASSP)}, pages 7335--7339.

\bibitem[{Vovk(2012)}]{vovk2012conditional}
Vladimir Vovk. 2012.
\newblock \href {https://proceedings.mlr.press/v25/vovk12.html} {Conditional
  validity of inductive conformal predictors}.
\newblock In \emph{Proceedings of the Asian Conference on Machine Learning},
  volume~25 of \emph{Proceedings of Machine Learning Research}, pages 475--490,
  Singapore Management University, Singapore. PMLR.

\bibitem[{Vovk(2015)}]{vovk2012crossconformal}
Vladimir Vovk. 2015.
\newblock \href {https://doi.org/10.1007/s10472-013-9368-4} {Cross-conformal
  predictors}.
\newblock \emph{Annals of Mathematics and Artificial Intelligence},
  74(1):9--28.

\bibitem[{Vovk et~al.(2005)Vovk, Gammerman, and Shafer}]{vovk2005CP}
Vladimir Vovk, Alex Gammerman, and Glenn Shafer. 2005.
\newblock \emph{Algorithmic Learning in a Random World}.
\newblock Springer-Verlag, Berlin, Heidelberg.

\bibitem[{Vovk et~al.(2018)Vovk, Nouretdinov, Manokhin, and
  Gammerman}]{pmlr-v91-vovk18a}
Vladimir Vovk, Ilia Nouretdinov, Valery Manokhin, and Alexander Gammerman.
  2018.
\newblock \href {https://proceedings.mlr.press/v91/vovk18a.html}
  {Cross-conformal predictive distributions}.
\newblock In \emph{Proceedings of the Seventh Workshop on Conformal and
  Probabilistic Prediction and Applications}, volume~91 of \emph{Proceedings of
  Machine Learning Research}, pages 37--51. PMLR.

\bibitem[{Vovk and Petej(2014)}]{vovk2014vennabers}
Vladimir Vovk and Ivan Petej. 2014.
\newblock Venn-abers predictors.
\newblock In \emph{Proceedings of the Thirtieth Conference on Uncertainty in
  Artificial Intelligence}, UAI'14, page 829–838, Arlington, Virginia, USA.
  AUAI Press.

\bibitem[{Wang et~al.(2024)Wang, He, and Kantaros}]{wang2024safe}
Jun Wang, Guocheng He, and Yiannis Kantaros. 2024.
\newblock \href {https://arxiv.org/abs/2402.15368} {Safe task planning for
  language-instructed multi-robot systems using conformal prediction}.
\newblock \emph{arXiv preprint arXiv:2402.15368}.

\bibitem[{Wang et~al.(2021)Wang, Choi, Xu, and Yang}]{wang2021putting}
Zijie~J. Wang, Dongjin Choi, Shenyu Xu, and Diyi Yang. 2021.
\newblock \href {https://aclanthology.org/2021.hcinlp-1.8} {Putting humans in
  the natural language processing loop: A survey}.
\newblock In \emph{Proceedings of the First Workshop on Bridging
  Human{--}Computer Interaction and Natural Language Processing}, pages 47--52,
  Online. Association for Computational Linguistics.

\bibitem[{Wiegreffe and Pinter(2019)}]{wiegreffe-pinter-2019-attention}
Sarah Wiegreffe and Yuval Pinter. 2019.
\newblock \href {https://doi.org/10.18653/v1/D19-1002} {Attention is not not
  explanation}.
\newblock In \emph{Proceedings of the 2019 Conference on Empirical Methods in
  Natural Language Processing and the 9th International Joint Conference on
  Natural Language Processing (EMNLP-IJCNLP)}, pages 11--20, Hong Kong, China.
  Association for Computational Linguistics.

\bibitem[{Xiao and Wang(2019)}]{Xiao_Wang_2019}
Yijun Xiao and William~Yang Wang. 2019.
\newblock \href {https://doi.org/10.1609/aaai.v33i01.33017322} {Quantifying
  uncertainties in natural language processing tasks}.
\newblock \emph{Proceedings of the AAAI Conference on Artificial Intelligence},
  33(01):7322--7329.

\bibitem[{Xu and Xie(2021)}]{xu2021conformal}
Chen Xu and Yao Xie. 2021.
\newblock \href {https://proceedings.mlr.press/v139/xu21h.html} {Conformal
  prediction interval for dynamic time-series}.
\newblock In \emph{Proceedings of the 38th International Conference on Machine
  Learning}, volume 139 of \emph{Proceedings of Machine Learning Research},
  pages 11559--11569. PMLR.

\bibitem[{Ye et~al.(2024)Ye, Yang, Pang, Wang, Wong, Yilmaz, Shi, and
  Tu}]{ye2024benchmarking}
Fanghua Ye, Mingming Yang, Jianhui Pang, Longyue Wang, Derek~F. Wong, Emine
  Yilmaz, Shuming Shi, and Zhaopeng Tu. 2024.
\newblock \href {https://arxiv.org/abs/2401.12794} {Benchmarking llms via
  uncertainty quantification}.
\newblock \emph{arXiv preprint arXiv:2401.12794}.

\bibitem[{Zerva et~al.(2022)Zerva, Glushkova, Rei, and
  Martins}]{zerva-etal-2022-disentangling}
Chrysoula Zerva, Taisiya Glushkova, Ricardo Rei, and Andr{\'e} F.~T. Martins.
  2022.
\newblock \href {https://doi.org/10.18653/v1/2022.emnlp-main.591}
  {Disentangling uncertainty in machine translation evaluation}.
\newblock In \emph{Proceedings of the 2022 Conference on Empirical Methods in
  Natural Language Processing}, pages 8622--8641, Abu Dhabi, United Arab
  Emirates. Association for Computational Linguistics.

\bibitem[{Zerva and Martins(2023)}]{zerva2023conformalizing}
Chrysoula Zerva and Andr{\'e}~FT Martins. 2023.
\newblock \href {https://arxiv.org/abs/2306.06221} {Conformalizing machine
  translation evaluation}.
\newblock \emph{arXiv preprint arXiv:2306.06221}.

\bibitem[{Zhan et~al.(2022)Zhan, Wang, and Gevaert}]{zhan2022reliable}
Xianghao Zhan, Fanjin Wang, and Olivier Gevaert. 2022.
\newblock \href {https://doi.org/10.1109/JBHI.2022.3193365} {Reliably filter
  drug-induced liver injury literature with natural language processing and
  conformal prediction}.
\newblock \emph{IEEE Journal of Biomedical and Health Informatics}, PP:1--9.

\bibitem[{Zhan et~al.(2023)Zhan, Xu, Zheng, Lu, and
  Gevaert}]{zhan2023reliabilitybased}
Xianghao Zhan, Qinmei Xu, Yuanning Zheng, Guangming Lu, and Olivier Gevaert.
  2023.
\newblock \href {https://arxiv.org/abs/2309.07332} {Reliability-based cleaning
  of noisy training labels with inductive conformal prediction in multi-modal
  biomedical data mining}.
\newblock \emph{arXiv preprint arXiv:2309.07332}.

\bibitem[{Zhao et~al.(2024)Zhao, Chen, Yang, Liu, Deng, Cai, Wang, Yin, and
  Du}]{zhao2023explainability}
Haiyan Zhao, Hanjie Chen, Fan Yang, Ninghao Liu, Huiqi Deng, Hengyi Cai,
  Shuaiqiang Wang, Dawei Yin, and Mengnan Du. 2024.
\newblock \href {https://doi.org/10.1145/3639372} {Explainability for large
  language models: A survey}.
\newblock \emph{ACM Trans. Intell. Syst. Technol.}, 15(2).

\bibitem[{Zollo et~al.(2023)Zollo, Morrill, Deng, Snell, Pitassi, and
  Zemel}]{zollo2023prompt}
Thomas Zollo, Todd Morrill, Zhun Deng, Jake Snell, Toniann Pitassi, and Richard
  Zemel. 2023.
\newblock \href {https://openreview.net/forum?id=zrgqy65sQw} {Prompt risk
  control: A rigorous framework for responsible deployment of large language
  models}.
\newblock In \emph{Socially Responsible Language Modelling Research}.

\end{thebibliography}
\bibliographystyle{acl_natbib}

\end{document}